\documentclass[preprint,5p,times,twocolumn]{elsarticle}

\usepackage{lineno}
\journal{Journal of Robotics and Autonomous Systems}

\usepackage{graphics} 
\usepackage{epsfig} 
\usepackage{mathptmx} 
\usepackage{times} 
\usepackage{amsmath} 
\DeclareMathOperator*{\argmax}{argmax} 
\usepackage{amssymb}  
\usepackage{xcolor}
\usepackage{algorithm2e}
\usepackage{soul}
\DeclareMathAlphabet{\mathcal}{OMS}{cmsy}{m}{n}

\makeatletter
\renewcommand{\@algocf@capt@plain}{above}
\renewcommand{\algocf@caption@plain}{\box\algocf@capbox\vskip\AlCapSkip}%
\makeatother

\usepackage[utf8]{inputenc}
\usepackage[pdfauthor={Bonetto et al.},
            pdftitle={iRotate:  Active  Visual  SLAM  for  Omnidirectional  Robots},
            pdfsubject={Active V-SLAM},
            pdfkeywords={View Planning for SLAM, Vision-Based Navigation, SLAM}]{hyperref}
\let\labelindent\relax
\usepackage{enumitem}
\usepackage{subfig}
\usepackage{cuted}
\usepackage{multirow}
\hypersetup{
    colorlinks=true,
    linkcolor=blue,
    filecolor=magenta,      
    urlcolor=blue,
}

\usepackage[font=footnotesize,labelfont=bf]{caption}
\usepackage{tikz}

\usepackage[textsize=tiny]{todonotes}

\newcommand{\uproman}[1]{\uppercase\expandafter{\romannumeral#1}}

\newcommand{\fig}[1]{Fig.~\ref{#1}}
\newcommand{\tab}[1]{Tab.~\ref{#1}}
\newcommand{\secref}[1]{Section~\ref{#1}}

\begin{document}

\begin{frontmatter}

\title{iRotate: Active Visual SLAM for Omnidirectional Robots}

\author[1,2]{Elia Bonetto\corref{cor1}}
\ead{elia.bonetto@tuebingen.mpg.de}
\author[2]{Pascal Goldschmid}
\ead{pascal.goldscmid@ifr.uni-stuttgart.de}
\author[1]{Michael Pabst}
\ead{michael.pabst@tuebingen.mpg.de}
\author[1]{Michael J. Black}
\ead{black@tuebingen.mpg.de}
\author[2,1]{Aamir Ahmad}
\ead{aamir.ahmad@ifr.uni-stuttgart.de}

\address[1]{Max Planck Institute for Intelligent Systems, Max Planck Ring 4, 72070, Tübingen, Germany.}
\address[2]{Institute for Flight Mechanics and Controls, The Faculty of Aerospace Engineering and Geodesy,\\ University of Stuttgart, Pfaffenwaldring 27, 70569 Stuttgart, Germany.}%
\cortext[cor1]{Corresponding author}

\begin{abstract}
In this paper, we present an active visual SLAM approach for omnidirectional robots. The goal is to generate control commands that allow such a robot to simultaneously localize itself and map an unknown environment while maximizing the amount of information gained and consuming as low energy as possible. Leveraging the robot's independent translation and rotation control, we introduce a multi-layered approach for active V-SLAM. The top layer decides on informative goal locations and generates highly informative paths to them. The second and third layers actively re-plan and execute the path, exploiting the continuously updated map and local features information.
Moreover, we introduce two utility formulations to account for the presence of obstacles in the field of view and the robot's location. Through rigorous simulations, real robot experiments, and comparisons with state-of-the-art methods, we demonstrate that our approach achieves similar coverage results with lesser overall map entropy. This is obtained while keeping the traversed distance up to $39\%$ shorter than the other methods and \textit{without} increasing the wheels' total rotation amount. Code and implementation details are provided as open-source, and all the generated data is available on-line for consultation.
\end{abstract}
\end{frontmatter}	
	
	\section{INTRODUCTION}
\label{sec:intro}

Mobile robots that assist humans in everyday tasks are increasingly popular, in both workplaces and homes. Self-localization and environment mapping are the key building-block state-estimation functionalities of such robots, which enable robustness in higher-level task performance, such as navigation, manipulation, and interaction with humans. Most often, these robots operate in a previously unknown environment, e.g., in a person's house. To develop the aforementioned state-estimation functionalities, a popular approach is V-SLAM -- visual simultaneous localization and mapping using onboard RGB cameras~\cite{CadenaCarloneCarrilloLatifScaramuzzaNeiraReidLeonard2016}.

Many V-SLAM methods are passive processes, e.g.~\cite{labbe2019rtab,rosinol2020kimera}, in which the robots are required to follow external control inputs and use their sensors only to build the map. Active V-SLAM, on the other hand, refers to those V-SLAM methods in which a robot makes decisions `on-the-fly' regarding its movements based on the state and sensors' readings, in such a way that both the map estimate and the higher-level tasks can benefit. Most active SLAM algorithms consist of the following workflow~\cite{CadenaCarloneCarrilloLatifScaramuzzaNeiraReidLeonard2016} -- i) selection of candidate goals, ii) path generation (set of waypoints) for those goals, iii) computation of path utilities, and, iv) executing/tracking the maximum utility path. Waypoints may either be reachable robot positions or include both position and headings (pose).

The insight in this work is that, during the path execution, the robot can continuously and actively control its camera heading to maximize the environment coverage by leveraging its omnidirectional nature. Coverage includes both the observation of previously unseen areas of the map and map refinement. By performing coverage in an active way, the robot can substantially speed up the information acquisition and quickly lower the overall map uncertainty. Our hypothesis is that this can significantly reduce the total distance the robot has to travel to map the environment, as also evidenced through our rigorous experimental results. A shorter travel distance translates to lower energy consumption and, therefore, a higher autonomy time, which is a critical aspect in many applications. 

\begin{figure}
	\centering
	\includegraphics[width=0.42\textwidth]{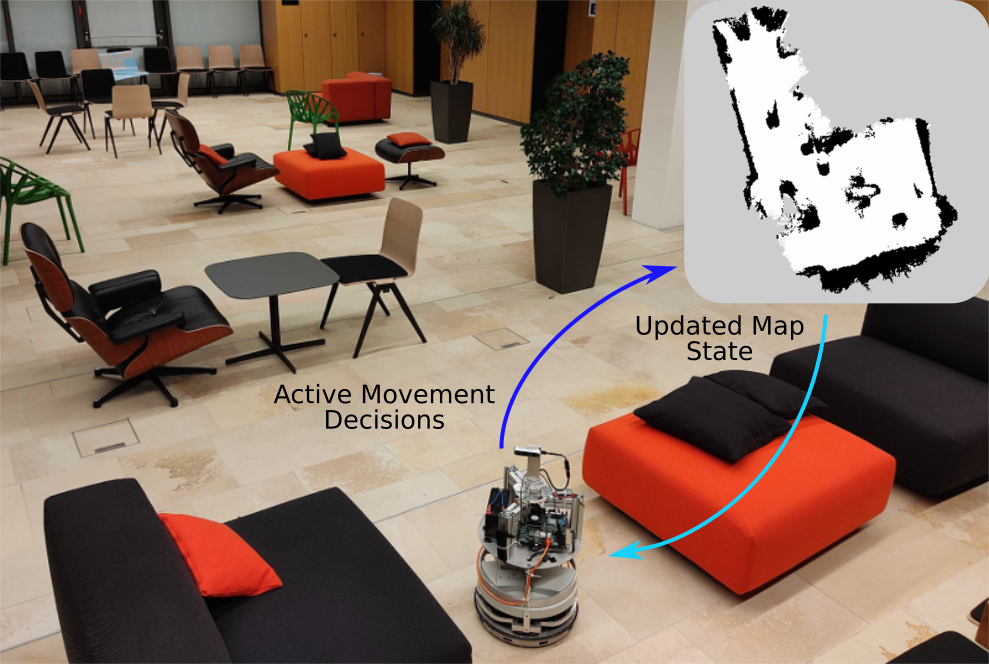}
	\caption{Our omnidirectional robot actively mapping an office reception area.}
	\label{fig:cover}
\end{figure}

Based on the above insight, the novel Active V-SLAM method we propose in this paper is made of three combined layers of activeness.

\begin{itemize}[leftmargin=*]

\item The paths toward available goals (in our case a set of frontier points) are optimized to consider, for every waypoint along those, the optimal heading direction to maximize a utility function. In this, we also account for the frustum overlap between waypoints. Therefore, we can exploit all the information contained in the map before performing any movement (see Sec.~\ref{subsec:globalpath}).
\item While executing the chosen optimal path, an intermediate refinement is done every time a waypoint is reached. In this, we re-compute the next waypoint's optimal heading based on the updated occupancy map. This allows us to account for dynamic variations in the map entropy and new information which, for example, loop closures or the robot's own movements might have brought between the time of reaching that waypoint and the time of the previous frontier point selection (see Sec.~\ref{subsubrefine}). 
\item Based on the previous step and on the real-time 3D features distribution we continuously refine the desired heading of the camera. In that way, we incorporate in the system the latest information acquired by the robot which was unavailable during both the previous steps. This, rather than being a visual odometry (VO) `helper', increases the visibility of the features to help loop closures events while the robot moves between waypoints (see Sec.~\ref{subsubfeat}). 
\end{itemize}

Thus, our V-SLAM approach is active throughout the mapping process rather than at fixed intervals of goals selection.

Furthermore, we also introduce (and experimentally compare) variations in the utility function based on the Shannon entropy of the map (Sec.~\ref{subsec:utility}). First, we use a weighted average between waypoints' utilities instead of a classic weighted sum. Second, we explicitly account for the presence of obstacles in the FOV and, third, propose a balance between the exploration and reobservation behaviours of the robot. We present our results and comparisons both in simulation and on a real robot. The latter is a Festo Didactic's Robotino, operating in an office environment (Fig.~\ref{fig:cover} and \ref{img:robotschmema}). It has an omnidirectional drive system and an onboard RGB-D camera, in addition to onboard computational units. The omnidirectional drive allows us to continuously control the camera heading without being restricted by the (non)holonomic constraints of other kind of platforms. Our method's source code and instructions for use are available open-source\footnote{\url{https://github.com/eliabntt/active_v_slam}}. All the generated data and the logging of each experiments is available for download\footnote{\url{https://keeper.mpdl.mpg.de/d/89f292ac267247df826f/}}.

The rest of the paper is structured as follows. In Sec.~\ref{sec:soa} we review the current state-of-the-art in active SLAM. Our method is described in Sec.~\ref{sec:approach}, while the experimental results are presented in Sec.~\ref{sec:exp}. In Sec.~\ref{sec:conc} we conclude with comments on future work.
	\section{Related Work}
\label{sec:soa}

Active SLAM algorithms address the exploration/exploitation dilemma in SLAM~\cite{CarrilloDamesKumarCastellanos2017,survey2021,sim2005global}, i.e., finding a balance between exploring new regions and exploiting previously acquired information to reduce overall map uncertainty.

Candidate goals are usually found alongside frontiers~\cite{CarrilloDamesKumarCastellanos2017,yamauchi1997frontier,umari2017autonomous,Dai2020}. Other approaches, like~\cite{Schmid2020} and~\cite{vallve2015active}, rely on persistent RRT-like structures, while in~\cite{chen2020active,LeungAttractorBased} the robot chooses the next set of actions based on the current global state. 
The goal with the highest associated utility is then selected. Computing that for all available paths is usually computationally expensive and, as shown in~\cite{stachniss2005information, CadenaCarloneCarrilloLatifScaramuzzaNeiraReidLeonard2016}, an accurate method to simulate uncertainty evolution and loop closures remains to be found. Therefore, approximations need to be introduced like in~\cite{CarrilloDamesKumarCastellanos2017,vallve2015active,rrt-uncertainty-aware}. In our approach, we overcome the issue of simulating these by i) planning and choosing the most informative global path using just the available map information, ii) discounting utility based on the length of the path to account for uncertainty, and iii) executing that path by continuous local re-planning.

The utility of a path is usually computed as a linear combination of several metrics~\cite{CadenaCarloneCarrilloLatifScaramuzzaNeiraReidLeonard2016} applied to the path's waypoints. These metrics usually comprise consistency of the map, path entropy, and robot state's uncertainty. The balance between exploration and exploitation is either performed implicitly within the utility formulation~\cite{CarrilloDamesKumarCastellanos2017} or based on some hand-crafted heuristics and threshold values~\cite{LeungAttractorBased,chen2020active}. The waypoint can be taken as-is or optimized based on the utility function itself. This is done either by using random sampling as in~\cite{rrt-uncertainty-aware}, or by checking the optimal orientation for each candidate pose~\cite{Schmid2020}. This can be combined with long-term planning in which the optimization of the pose is done a-priori, as in~\cite{Dai2020}, or in a receding horizon (RH) fashion where the chosen location will be reached immediately as in RH-NBV (next best view) planners~\cite{rrt-uncertainty-aware,Schmid2020}. In the latter case, the risk of getting stuck is high if the system is not carefully designed~\cite{survey2021,8633925} due to the local scope of the approach. Moreover, in an indoor scenario where usually 2D navigation is used, loop closures might degrade and drastically change the information of the map. Thus, blindly reaching the selected goal might be inefficient.
In our approach we combine both. All the waypoints along a path are pre-optimized and the utility of that sequence is computed using the distance between that point and the starting position as a discount factor. Then, during execution, we apply a refinement for every subsequent waypoint based on the utility function to ensure a constant adaptation to the newly obtained information (the first and second active levels mentioned in Sec.~\ref{sec:intro}). Therefore, we are able to combine the benefits of NBV planners (path heading optimization), RH techniques (next viewpoint optimization), and frontier-based exploration (long-term goal).

To compute each goal's utility, we not only consider the final waypoint, like in~\cite{Dai2020,umari2017autonomous}, but for the contribution of each waypoint on the path as in~\cite{vallve2015active,rrt-uncertainty-aware}. Nonetheless, using a plain weighted sum could not fully enclose the whole path contribution. Thus, we propose a weighted average approach that helps to capture how every \textit{step} of the path could contribute on average. 

Different utility functions have been proposed in the literature using Shannon~\cite{Dai2020} and Rényi~\cite{CarrilloDamesKumarCastellanos2017} entropy, exploration volume~\cite{7487281,8633925} or the quality of observations~\cite{Schmid2020}. The utility used in this work is based on the Shannon entropy, but we also propose and test two other ones based on obstacle presence in the FOV and on the position of the robot w.r.t. the final goal. 

Recently, learning applied to SLAM has gained momentum. Several methods applying deep learning and deep reinforcement learning emerged like~\cite{Chaplot2020Learning,Borui2019Active,WEN201529}. Some of those apply learning to both the SLAM backend and to the active SLAM procedure itself~\cite{Chaplot2020Learning}. Others, like~\cite{Borui2019Active,WEN201529}, instead build on top of already deployed SLAM frameworks, e.g. EKF-SLAM. However, works based in RL approaches usually limit the applicability to different robotic platforms. In general, comparing to these approaches is hard. Every RL agent is trained over specific traits including, but not limited to, platform constraints (e.g. in the velocity space) and different sensor sets and ranges (e.g. using 2D lidar for mapping). Moreover, changing the SLAM framework responsible to build the map might harm the system itself due to the different distribution of the extracted information. Therefore, a comparison with these methods would be unfair since any observed difference in the performance would not be linked to the method itself but possibly to many factors (e.g. learning procedure, constraints, control parameters).

Finally, real-time features and landmark following have been widely studied since the seminal work of Davison et al.~\cite{davison1998Mobile}. Tracking those elements is usually used to help visual odometry systems and/or direct localization. This is enforced either through an external independent component as in~\cite{4636757} or directly in the optimization method as in~\cite{LeungAttractorBased}. In contrast, we do not \textit{directly} consider the effect of keeping the features within the field of view, nor do we use previously observed features. 
In this work, we influence the viewpoint orientation based on the \textit{latest} features distribution and the computed optimal heading \textit{only} between waypoints. Our insight is that we can increase the robustness of the system `blindly' by balancing both the feature tracking and the information gain within the system --- all in real-time.
	\section{Proposed Approach}
\label{sec:approach}
\subsection{Problem Description and Notations}

Let there be an omnidirectional robot, traversing on a 2D plane in a 3D environment, whose pose is given by $\mathbf{x}_R = [x_R ~~ y_R ~~ \theta_R]^{\top}$ and velocity by $\mathbf{\dot{x}}_R = [\dot{x}_{R} ~~ \dot{y}_{R} ~~ \dot{\theta}_{R}]^{\top}$. Let the robot be equipped with an RGB-D camera with an associated maximum sensing distance $d_{thr}$ and horizontal field of view (FOV) $\alpha$, a 2D laser range finder, and an IMU.
\
Let $M \subset \mathbb{R}^2$ represent a bounded 2D grid map of the environment, where for each point $\mathbf{m} = [x_m ~~ y_m]^{\top} \in M$ we have an occupancy probability $p_o(\mathbf{m}) \in [0,1]$. 
\
We assume that the robot begins exploration from a collision-free state and all map cells belong to the \textit{unknown} space $M_{unk} \subset M$. 
\
The robot's goal is to quickly, efficiently (i.e, with minimal energy consumption or travelled distance), and autonomously map all the observable points in $M_{unk}$ as \textit{free} ($M_{free}$) or \textit{occupied} ($M_{occ}$). The problem is considered solved when the $M_{occ} \cup M_{free} \cup M_{hid} = M$ is verified, where $M_{hid}$ represents the space that cannot be mapped as, for example, unreachable locations and $M_{free} \cup M_{occ} = M_{exp}$ represent the already explored space. Some cells of this map are further identified as frontiers. These are cells located at the boundaries between \textit{free} and \textit{unknown} spaces, which are large enough for the robot to be traversed. The robot must also keep the map's uncertainty as low as possible while exploring and have a final overall low trajectory error.

At any given time instant the robot's state, map state, the set of observed visual features, and the graph $\mathbf{G}$ of previously visited locations generated by the V-SLAM backend are available to the robot. A node $\mathbf{n} = [x ~~ y ~~ \theta]^{\top} \in \mathbf{G}$ is defined by its coordinates in the map frame and by the orientation of the robot. While solving this problem the robot must also be able to avoid both static and dynamic obstacles. Global and local path planning have to be performed without any knowledge of the environment or any landmark in it. 

We propose an active SLAM approach to solve the above-described mapping problem. The crux of our approach is to explore the environment by generating and executing paths that maximize information acquisition by the robot's sensors. More precisely, at a given instant, the robot i) identifies a suitable exploration goal position given by $\mathbf{x}_G = [x_G ~~ y_G ~~ \theta_G]^{\top}$, ii) based on the so-far acquired information up to that instant, generates a collision-free path, which consists of contiguous poses $\mathbf{\sigma} = \{\mathbf{x}_R, ..., \mathbf{x}_G\}$, with maximum possible future information gain, and iii) executes the path by continuously re-planning its heading direction to incorporate all acquired information along the way and keep in its FOV as many visual features as possible. We leverage the omnidirectional nature of the robot platform, i.e., the position and orientation of the robot can be controlled independently. The exploration is considered concluded after a certain amount of time has expired since the robot is unaware of the total amount of area that has to be explored.
\
In this work, we use Shannon entropy as a measure of the map's uncertainty. The map entropy is therefore given by 

\begin{small}
\begin{align}
    E_{\textrm{map}} = \sum_{\mathbf{m} \in M_{exp}} {E[\mathbf{m}]} = & \sum_{\mathbf{m} \in M_{exp}}  -(p_o(\mathbf{m})\cdot \log_2(p_o(\mathbf{m})) \\ \nonumber &+ (1-p_o(\mathbf{m}))\cdot \log_2(1-p_o(\mathbf{m}))),
    \label{eq:shannon}
\end{align}\end{small}where $E[\mathbf{m}]$ is the Shannon entropy of a map cell $\mathbf{m}$ and $p_o(\mathbf{m})$ represents its occupancy probability. $p_o(\mathbf{m}) = 0.5$ if $\mathbf{m} \in M_{unk}$. 

We further describe our goal selection approach in Sec.~\ref{subsec:goalsel}. The path's utilities formulation based on Shannon entropy are introduced in Sec.~\ref{subsec:utility}. Our first active level, i.e. the global path planning, is described in Sec.~\ref{subsec:globalpath}. The second and third levels of activeness are introduced in Sec.~\ref{subsec:refine}. More precisely, the path execution and local re-planning can be found respectively in Sec.~\ref{subsubrefine} and Sec.~\ref{subsubfeat}. Finally, in Sec.~\ref{subsec:FSM}, we outline a finite state machine necessary to coordinate the proposed active SLAM approach.

\subsection{Goal selection}
\label{subsec:goalsel}

Reachable frontier map cells are defined as those which have no obstacles in the neighbouring cells and belong to a region that is big enough for the robot to traverse. We use the frontier extraction algorithm from~\cite{Mammolo2019}, and modify it in the following way. 
\
At first, frontier clusters are identified. Next, candidate goal locations are selected at centroids of those clusters. If a centroid is not reachable, a greedy search is performed in its proximity to identify the next reachable frontier cell in that cluster. These are then marked as candidate goals. Further, if any of the candidate goals was the selected goal of the previous iteration, it is discarded assuming that it was unreachable. This is followed by the process of global path generation (Sec.~\ref{subsec:globalpath}) to the candidate goals and utility computation of those paths (Sec.~\ref{subsec:utility}). The candidate goal with the maximum utility path is then chosen as the actual goal position.
\
Differently to the most common approach for frontier-based autonomous exploration, we do not consider the task complete if frontiers are not available~\cite{CarrilloDamesKumarCastellanos2017,survey2021,yamauchi1997frontier,Dai2020}. In such cases, our robot either does a complete in-place rotation or tries to move to a randomly-picked previously-visited location, until the allotted time for the experiment has elapsed. This allows the robot to refine the map, activate new loop closures, and, eventually, find out new frontiers. The robot will reach the selected goal with newly generated paths, i.e. not necessarily tracking back its previous movements, while still actively mapping the environment using all the activeness levels presented later. This selection is completely random as we do not search for high-entropy, feature-rich areas, or known old loop closures locations. We consider this a good trade-off between stopping the exploration and manually selecting a goal. The described strategy is similar to what is employed in~\cite{stachniss2005information}. However, the work of Stachniss et al. presents several differences w.r.t. ours, e.g. they consider these alternative actions \textit{while} performing explorations, believe the map uncertainty minimal when no unknown areas are left, and terminate the experiment as soon as no gain is expected in the pose uncertainty. Differently, we recognize that the uncertainty is not given to be minimal in such situations, and new frontiers might be found after, for example, a loop closure or a refinement of a map portion.
Moreover, forcibly selecting previously-found loop closures might result also in an overall drifting or falling into local optima if not carefully tuned~\cite{CadenaCarloneCarrilloLatifScaramuzzaNeiraReidLeonard2016}. Another approach might be selecting only high entropy areas. However, in such cases, the system might suffer excessive drift or not even being able to reach the goal locations at all, e.g. when placed in a tight spot. Moreover, since we try to reduce the entropy of the map by continuously controlling the heading of the robot with our three active levels, high-entropy areas might be directly covered while following the random objective.
Finally, differently than~\cite{LeungAttractorBased} and similar related works, we do not have any other behaviour implemented that acts \textit{concurrently} to the exploration to which we can resort, nor we have a underlying RRT representation like~\cite{8633925,rrt-uncertainty-aware}. However, while the latter suffer from the necessity of carefully tuned parameters to successfully terminate the exploration~\cite{8633925}, the former is based on maintaining a global set of features which is not our case.


\subsection{Utilities}
\label{subsec:utility}

We use the utility as a measure of information gain for an attainable robot pose $\mathbf{x}$ (position and orientation) on the map that lies along a path $\mathbf{\sigma}$, leading to the goal location $\mathbf{x}_G$. Given a pose $\mathbf{x}$, we obtain a set of $N$ map cells $C_{\mathbf{x}} = \{\mathbf{m}_0, ..., \mathbf{m}_N\}$ that are visible from the pose $\mathbf{x}$. A cell $\mathbf{m}$ is deemed visible if there exists a ray that goes from it to the queried pose $\mathbf{x}$ without intercepting any obstacle and lies within the robot camera's FOV, hypothesizing the robot at pose $\mathbf{x}$. A cell is considered an obstacle if its $p_o(\mathbf{m})>0.7$. At every cell $\mathbf{m} \in {C_{\mathbf{x}}}$, we associate a utility value $u[\mathbf{m}|\mathbf{x},\mathbf{\sigma}]$.
\
The total utility $U_{[\mathbf{x},\mathbf{\sigma}]}$ of a pose $\mathbf{x}$ along a path $\mathbf{\sigma}$, is given as 

\begin{small}
\begin{equation}
 U_{[\mathbf{x},\mathbf{\sigma}]} = \sum_{\mathbf{m} \in \mathbf{C_{\mathbf{x}}}} u[\mathbf{m} | \mathbf{x},\mathbf{\sigma}].
 \label{eq:locationutility}
\end{equation}
\end{small}

We compare three different ways to compute $u[\mathbf{m} | \mathbf{x},\mathbf{\sigma}]$. The first way, $u_1$, is to define it as the plain Shannon entropy, which represents the available amount of information, i.e., $u_1[\mathbf{m}| \mathbf{x},\mathbf{\sigma}] = E[\mathbf{m}]$. With this, we can capture directly the maximum amount of information retrievable from a given cell. In the second way, $u_2$, we introduce the presence of obstacles directly in the utility computation.

\begin{small}
\begin{equation}
    u_2[\mathbf{m}| \mathbf{x},\mathbf{\sigma}] = \begin{cases}
    E[\mathbf{m}] & \text{ if } p_o(\mathbf{m}) < p_{thr} \\ 
    E[\mathbf{m}] + \kappa_1 & \text{ if } p_o(\mathbf{m}) \geq p_{thr} ~~ ,
    \end{cases}
\end{equation}\end{small}where $\kappa_1$ is a constant (set to $1$ in our experiments), and $p_{thr}$ is the probability of a cell being an obstacle, set to $0.7$ in our experiments, i.e. the probability of being an obstacle. The insight is twofold in this case. First, we know that features usually are found nearby obstacle regions where we can observe contrast and variability against the background. At the same time, those are harder to be mapped correctly with respect to empty areas. However, the entropy of those cells goes rapidly toward 0 due to Shannon’s formulation. Therefore, adding this constant term to account for obstacle presence will favour robot headings towards that kind of cells improving the likelihood of observing more features and `forcing' it to observe objects in order to refine their mapping. Differently from $u_1$, here we \textit{explicitly} favour the observation of obstacle cells.

The third way of computing utility, $u_3$, is based on the following idea. Along the path, a robot can balance the observation of new, unmapped cells and the re-observation of previously mapped ones. Generally, if a robot is far w.r.t. a frontier point the amount of unknown area it can explore locally is very limited but has a big impact on the utility function. This is due to the fact that unmapped cells have $p_o(\mathbf{m})=0.5$, therefore their utility is equivalent to $E[\mathbf{m}] = 1$. With $u_3$ we aim to emphasize the refinement of already mapped cells rather than exploring the new area while the robot is far w.r.t. the final goal and vice-versa. With this, we seek to differentiate the objective dynamically between re-observation (refinement) and exploration, by differently weighting the two based on the distance between the robot and the frontier point. To this end, we first define 
\begin{small}
\begin{equation}
\lambda_{[\mathbf{x},\mathbf{\sigma}]}[\mathbf{m}] = \begin{cases}
    (1-j[\mathbf{x},\mathbf{\sigma}]) & \text{ if } \mathbf{m} \in M_{unk} \\ 
    j[\mathbf{x},\mathbf{\sigma}] & \text{ if } \mathbf{m} \in M_{free} \bigcup M_{occ} ~~,
\end{cases}
\end{equation}\end{small}as our weighting function, where $j$ itself is a function of the distance between the queried pose $\mathbf{x}$ and the final goal pose $\mathbf{x}_G$ given as
\begin{equation}
    j = \max(d_l, \min(d_h,  d_l||\mathbf{x}-\mathbf{x}_G||_2)).
\label{eq:j}
\end{equation}
\
Constants $d_l$ and $d_h$ are parameters to tune re-observation priorities (set to $0.2$ and $0.8$, respectively, in our experiments). Note that it is always verified that $j \in [0.2, 0.8]$ and the further we are from $\mathbf{x}_G$ the nearer we are to $d_h$. Using this weighting function, $u_3$ is given as

\begin{small}
\begin{equation}
    u_3[\mathbf{m} | \mathbf{x},\mathbf{\sigma}] = \begin{cases}
    \lambda_{[\mathbf{x},\mathbf{\sigma}]}[\mathbf{m}] E[\mathbf{m}]   & \text{ if } p_o(\mathbf{m}) < p_{thr} \\ 
    \lambda_{[\mathbf{x},\mathbf{\sigma}]}[\mathbf{m}] E[\mathbf{m}] + \kappa_1  & \text{ if } p_o(\mathbf{m}) \geq p_{thr}. \end{cases}
\end{equation}\end{small}

By using $u_3$ we increase the weight of the utility related to the entropy of the already explored cells while the robot is far from the final frontier point. Then, while the robot moves toward the exploration zone (the frontier point) the weight is shifted toward the utility related to the unmapped cells. We built this on top of $u_2$, meaning that we keep the obstacle weight for the reasons explained above. However, different from both $u_1$ and $u_2$, here we dynamically balance between reobserving the already explored area and mapping new space.

\subsection{Global path generation -- \textit{\textbf{first-level activeness}}}
\label{subsec:globalpath}

For every available candidate goal, we generate a path with an A$^\star$ planner applied to the global occupancy grid. Every path $\mathbf{\sigma}$ obtained in this way is reduced to a set of waypoints $W = \{\mathbf{w}_0, ... , \mathbf{w}_G\}$, where a waypoint is defined as $\mathbf{w}_i = [x_i ~~ y_i ~~ \theta_i]^{\top}$, $\mathbf{w}_0 = \mathbf{x}_R$, $\mathbf{w}_G = \mathbf{x}_G$. The path length between waypoints is restricted to a fixed value ($1 \mathrm{m}$ in our experiments). This discretization is done for several reasons. First, this lowers the overall computational requirement of the complete active SLAM approach. Second, it facilitates the robot's dynamic constraints, e.g., it allows the robot sufficient time to rotate such that the required heading of the subsequent waypoint is achievable and unnecessary in-place rotations are avoided. 

While the A* planner also returns a heading for every waypoint based on some heuristics, a naive but non-informative approach would be to interpolate the A* trajectory following movement direction in order to obtain the headings. However, this limits the observation capabilities of the system. Since an omnidirectional platform can perform instantaneous roto-translational movements we can leverage this characteristic to enrich the path with more informative view orientations. In this way, instead of having a series of mostly overlapping interpolated headings along the path, we obtain a different and enhanced series of robot's orientations. Those are computed such that the utility of the path is optimized with respect to the information that is currently available. Note that, differently to the case with non-omnidirectional platforms, in this situation, we do not need to recompute the approach path since the robot has no limitations in its movement capabilities. Therefore, for our \textit{first-level activeness}, we compute the optimal heading for every waypoint. In this, for every waypoint $\mathbf{w}_i = [x_i ~~ y_i ~~ \theta_i]^{\top}$ we replace $\theta_i$ by $\theta^\star$, where the latter maximizes the waypoint's utility value as follows

\begin{equation}
 \theta^\star =  \argmax_{\theta_i \in [0,2\pi]} U_{[\mathbf{w}_i,\mathbf{\sigma}]}.
\end{equation}
However, instead of directly using (\ref{eq:locationutility}) to compute $U_{[\mathbf{w}_i,\mathbf{\sigma}]}$, we update the map cell set ${C_{\mathbf{w}_i}}$ in a way such that the cells visible from the waypoints prior to $\mathbf{w}_i$ are not included in $C_{\mathbf{w}_i}$, i.e. we account for the frustum overlap of the subsequent camera views. The final path utility is then given by 

\begin{equation}
    {U}[\mathbf{\sigma}] = \frac{\sum_{i = 0}^{N} {k_i \cdot U_{[\mathbf{w}_i,\mathbf{\sigma}]}}}{\sum_{i = 0}^{N}{k_i}},
\end{equation}where $k_i = e^{-\rho \cdot d_i}$ is a discount factor, $d_i$ is the path-distance between the waypoint $\mathbf{w}_i$ and the current robot location and $\rho$ is a discount factor set to $0.25$ in our experiments. Instead of the classical approach, like the one used in~\cite{rrt-uncertainty-aware} that considers a plain weighted sum, we apply a weighted average between all waypoints utilities. This choice enables us to capture a more comprehensive contribution of the whole path in terms of overall information gain. With a weighted sum the system favors paths that exhibit spikes in the utility, caring only about a high cumulative reward rather than how that reward is obtained and how much it will gain at every step. Instead, the weighted average approach considers how much the robot can gain along the \textit{whole} path seeking routes that are more informative on average irrespective of their length. Finally, the highest utility path is selected and its waypoints are used by the local path execution module, as described next.


\begin{figure*}[!ht]
	\centering

    \includegraphics[scale=.34]{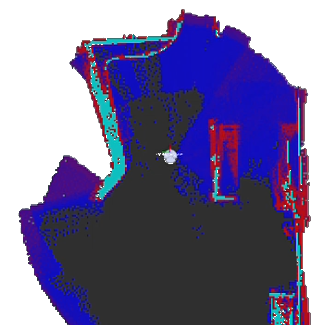}
    \includegraphics[scale=.34]{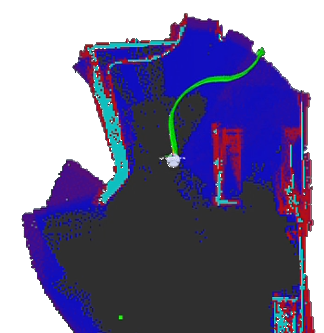}
    \includegraphics[scale=.34]{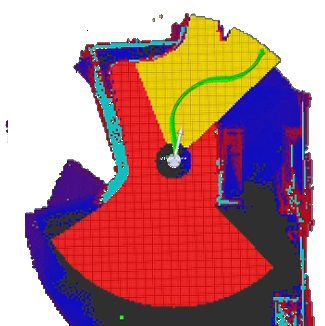}
    \includegraphics[scale=.2775]{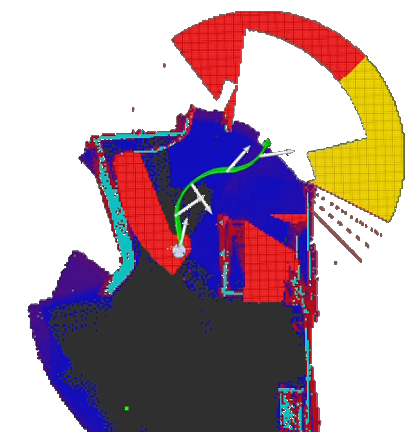}
    \includegraphics[scale=.314]{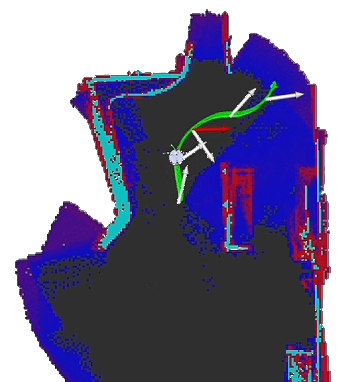}
    
    \caption{A representation of the first and second levels of activeness. The initial state, shown in the first figure, depicts the current voxels' occupancy probability based on RViz visualization, i.e. black represent free space, blue is likely free, red is likely an obstacle and light blue is obstacle. In the second figure, we can notice the frontier points (green dots) and the path generated toward one of them. For every waypoint along this path, we compute the optimal heading and the frustum overlap (3rd and 4th figures), i.e. our \textbf{\textit{first level of activeness}}. The red circular area is the set of visible cells from a given location. The yellow area represents the ones covered by the selected optimized orientation based on the utility function. Optimal headings computed by the first level of activeness are depicted as white arrows in figures 4 and 5 of the sequence. In figure 4 we can observe how many cells are not being considered due to the previous headings selection. In the final image we can observe the difference that exists between the pre-computed optimal view direction (white arrow) and the one computed once the previous waypoint is reached (red arrow), i.e. \textit{\textbf{second level of activeness}}. While reaching the re-optimized heading (i.e. the red arrow) the robot continuously controls its orientation based on the \textbf{\textit{third active level}}.}
    
	\label{img:methodschema}
\end{figure*}

\subsection{Local re-planning and execution}
\label{subsec:refine}

We use a nonlinear model predictive control (NMPC) based method for the local path execution. The NMPC formulation is based on the kinematic description of a three-wheeled omnidirectional robot. The model is nonlinear because of the simultaneous rototranslation capabilities of the omnidirectional platform. The kinematic model can be found, for example, in Sec. 3-B of~\cite{9568791}, which we omit here for brevity. The state $\mathbf{x}_t$ is the robot's position and orientation, while we control the robot in velocity using the vector $\mathbf{u}_t(n) = [u_{t,x} ~~ u_{t,y} ~~ u_{t,\theta}]^T$, where $n$ is the horizon time step. The objective is to reach the desired goal location, i.e. the next waypoint $\mathbf{w}$. The desired orientation is provided by the second and the third level of activeness (see Sec.~\ref{subsubrefine} and Sec.~\ref{subsubfeat}). In order to ensure safe operations the robot must be able to avoid obstacles and stay within its control bounds. Thus, the MPC is formulated as follows \begin{equation}
\mathbf{x}_t(0)^{\star} \dots \mathbf{x}_t(N)^{\star}, \mathbf{u}_t^{\star}(0) \dots \mathbf{u}_t^{\star} (N-1) 
= \underset{\mathbf{u}_t(0) \dots \mathbf{u}_t(N-1)}{\pmb{\arg\min}} (J)
\end{equation}
\begin{equation}
    \mathtt{s.t.}\ \mathbf{x}_t(n+1) = f(\mathbf{x}_t(n),\mathbf{u}_t(n))
\end{equation}
\begin{equation}
    \begin{split}
    &-\mathbf{u}_{max} \leq \mathbf{u}_t(n) \leq \mathbf{u}_{max}   \\
    &-{v}_{tr, max} \leq {v}_{t,tr}(n) \leq {v}_{tr, max}   \\    
    &{d}^k_{t}(n) \geq {d}_{min},\ \forall\ k\ \in \mathbf{obs}(n)   \\    
    \end{split}
\end{equation}
Where $f$ is the nonlinear kinematic equation (see Sec. 3-B in~\cite{9568791}), $v_{t,tr}(n)$ represents the \textit{overall} translation velocity, i.e. $(u_{t,x}^2 + u_{t,y}^2)^{0.5}$. Since the environment could be dynamic, we consider $\mathbf{obs}(n)$ to be the vector containing all the predicted obstacle positions and $d^{k}_t(n)$ to be the distance between the robot and the \textit{k}-th obstacle at the $n$-th time step. We keep ${u}_{\{x,y,\theta\}, max}, v_{tr, max}, d_{min}\in \mathbb{R}^+$.

The robot must minimize the distance to the target, the control effort, and a cost for being nearby an obstacle. The objective $J$ can be defined as follows
\begin{equation}
    \begin{split}&J = (\mathbf{x}_t(N) -\mathbf{w})^T \mathcal{Q}_{x} (\mathbf{x}_t (N)-\mathbf{w}) + \\
    &+ \sum_{n=0}^{N-1}\left [\vphantom{\sum_n^k} (\mathbf{x}_{t}(n)-\mathbf{w})^T \mathbf{\mathcal{Q}}_{x} (\mathbf{x}_{t}(n)-\mathbf{w}) + {\mathbf{u}}_t(n)^T \mathcal{R} {\mathbf{u}}_t(n) + \right.\\
    &\left. {\mathbf{z}}_{t}(n)^T \mathcal{Q}_{obs} {\mathbf{z}}_{t}(n) \vphantom{\sum_n^k} \right ]
\end{split}
\end{equation}
${\mathbf{z}_t(n)} = [h^1_t(n) ~~  \dots ~~ h^k_t(n)]^T$ indicate a `force' factor that drives the robot away from the obstacles. This is computed, for each one of the $k$ obstacles, as $ h^k_t(n)= -(1 - e^{0.1 / d^{k}_t(n)^2})$. We add this term, in addition to the minimum distance constraint, to further discourage the robot to go nearby them in order to better cope with noise due to the SLAM procedure and possible errors in their localization. With $\mathcal{Q}_x$, $\mathcal{R}$, and $\mathcal{Q}_{obs}$ defined as the applied positive diagonal weighting matrices. The implementation details can be found in~\secref{setupandimplementation}.

Note that while A* provides an obstacle-free path, it is necessary to include obstacle constraints in the local re-planning and execution to avoid collisions with any newly discovered obstacle cell. We further describe the last two activeness levels, which ensure active information acquisition throughout the whole path execution.

\subsubsection{Local waypoint refinement -- \textit{\textbf{second-level activeness}}}
\label{subsubrefine}

The \textit{second-level activeness} is carried out immediately before sending a waypoint to the NMPC. 
\
Here, we exploit the fact that the robot has already moved and mapped new areas of the environment since the global path was computed.
\
The information considered during the previous global planning and optimal heading generation could have changed drastically, e.g., due to loop closures events, intermediate robot movements, newly mapped obstacles, or map corruptions. 
\
Therefore, each time the robot reaches a waypoint $\mathbf{w}_i$, our method re-computes the optimal heading of the subsequent waypoint $\mathbf{w}_{i+1}$, using the same procedure as described at the end of Subsec.~\ref{subsec:globalpath}, but by using the updated map information, obtaining a refined $\Tilde{\theta}^\star_{i+1}$ that will be used as objective. In this way, we ensure that we consider the most recent and updated version of the occupancy grid not only when we do the global long-term planning, but also every time we reach a sub-goal.

\subsubsection{Real-time heading refinement -- \textit{\textbf{third-level activeness}}}
\label{subsubfeat}
\begin{figure}[h]
    \centering
    \includegraphics[width=.9\columnwidth]{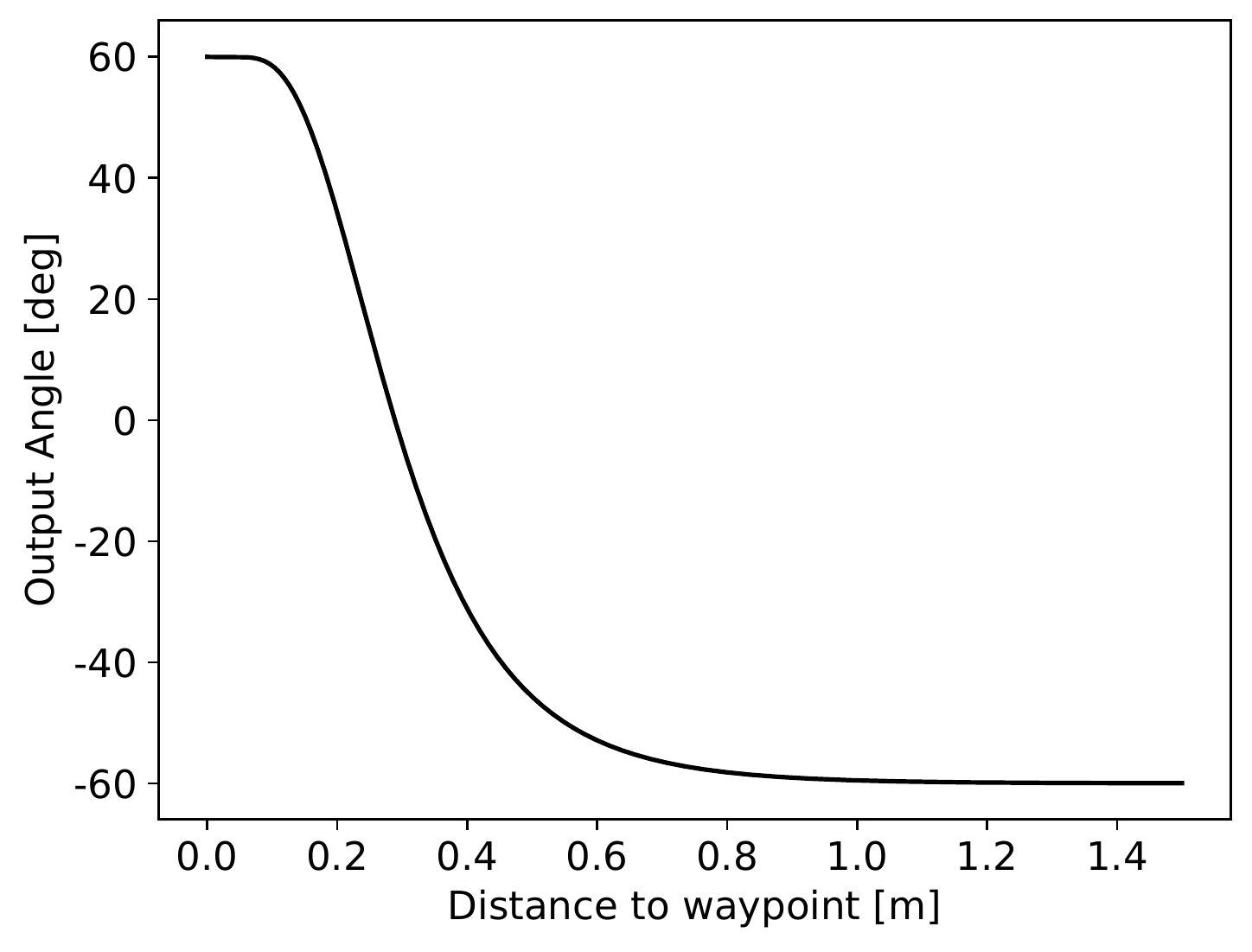}
    \caption{Computed $\gamma_t$ with $\tilde{\theta}^\star_{i+1} = 60 \mathtt{ deg}$, $\kappa_2 = -6$, $\beta_{i+1} = -60 \mathtt{ deg}$, $\kappa_3 = -0.5$, and $d_t \in [0,1.5]$.}
    \label{fig:weighting}
\end{figure}

The V-SLAM back-end computes a set of 3D features for every time step. In the \textit{third-level activeness}, which is carried out in \textit{between} two consecutive waypoints, the key idea is to keep in the robot camera's FOV as many 3D visual features as possible.
\
For every detection step in V-SLAM (i.e., node added to the graph), we obtain the 3D locations of all the features that are being extracted from the current image. Therefore, only the features relative to the latest added node are being used. We use this information to actively steer the heading of the camera. Note that this information is not available during the previous step as it is only computed in real-time. Therefore, the planning that was carried out during the first and the second active levels could not consider this. Moreover, even if a global feature map was to be maintained, new and better features might be found within the newly taken images.
\
At any given time instant $t$ while traversing between waypoints $\mathbf{w}_i$ and $\mathbf{w}_{i+1}$, we have i) $\Tilde{\theta}_{i+1}^\star$, the desired optimal angle of the next waypoint $\mathbf{w}_{i+1}$ as computed by the \textit{second-level activeness}, ii) $\beta_{i+1}$, the orientation that the robot should achieve to retain the maximum amount of features within its FOV from $\mathbf{w}_{i+1}$. At every time-step, we re-compute $\beta_{i+1}$ with a 3D ray-tracing technique on the updated octomap by using the horizontal FOV of the robot camera $\alpha$, the maximum distance of the features from it, and the next waypoint's 2D position. Let $d_t$ be the distance between the current robot position and the next waypoint $\mathbf{w}_{i+1}$. In real-time, we then compute the current desired orientation for the robot, $\gamma_t$,  as
\begin{equation}
    \gamma_t = \frac{\tilde{\theta}^\star_{i+1} \cdot e^{\kappa_2 \cdot d_t} + \beta_{i+1} \cdot e^{\kappa_3  \mathbin{/} d_t}}{e^{\kappa_2 \cdot d_t} + e^{\kappa_3  \mathbin{/} d_t}},
    \label{eq:balancefeat}
\end{equation}where $\kappa_2$ and $\kappa_3$ are constants. The NMPC's objective is now modified to track a continuously-updated waypoint $\mathbf{w}^{\prime}_{i+1} = [x_{i+1} ~~ y_{i+1} ~~ \gamma_t]^{\top}$, whose position component is the same as $\mathbf{w}_{i+1}$ but the orientation component is updated to $\gamma_t$. Varying the parameters $\kappa_2$ and $\kappa_3$ allows to vary the importance of keeping 3D visual features within the FOV versus achieving the previously-desired orientation $\Tilde{\theta}_{i+1}^\star$. For instance, in our experiments we set $\kappa_2 = -6$ and $\kappa_3 = -0.5$ to keep the features within the FOV as long as the robot is far enough w.r.t. the next waypoint. As it approaches the following waypoint the weight of the previously computed heading becomes more and more prominent. Indeed, since $\lim_{d_t \to 0^+} {e^{\kappa_2 \cdot d_t}} = 1$ and $\lim_{d_t \to 0^+} {e^{\kappa_3 \mathbin{/} d_t}} = 0$ we have that $\lim_{d_t \to 0^+}{\gamma_t} = \Tilde{\theta}_{i+1}^\star$. However, we acknowledge that features are not always available. In such scenarios the system sets $\beta_{i+1}$ to the \textit{current} heading direction of the robot to have anyway a smooth transition between the two following the same weighting procedure. A visual representation of this function with the chosen parameters is provided in~\fig{fig:weighting}
\
Overall, the \textit{third-level activeness} increases the chance of loop closures and place recognition (especially while revisiting areas), therefore improving map quality and lowering trajectory errors. 

\subsection{FSM}
\label{subsec:FSM}

A finite state machine coordinates the execution of the three active SLAM layers introduced in this work. A simplified scheme of the FSM is depicted in Alg.~\ref{alg:FSM}. The NMPC states (\textit{WAITING}, \textit{OPERATING}, \textit{GOAL REACHED}, or \textit{RECOVERY}) influence what the FSM does. If the NMPC is waiting for commands (\textit{WAITING, GOAL REACHED} states) the FSM will send a goal to it. If the current list of waypoints is empty, i.e. there is no goal already available, it will generate one. This is done by our first level of activeness and consists of both i) goal selection (see Sec.~\ref{subsec:goalsel}) and ii) global path generation (see Sec.~\ref{subsec:globalpath}). The optimized waypoints are then stored in a queue and executed one after the other. If instead the current list of waypoints is not empty the FSM will refine the heading of the next sub-goal, i.e. our second level of activeness (see Sec.~\ref{subsubrefine}), and send the optimized version to the NMPC. If the NMPC is actively working  (\textit{OPERATING}), the FSM will stay idle. Between each control loop, we take care of balancing the feature tracking and the current orientation objective via our third level of activeness (see Sec.~\ref{subsubfeat}). Specifically, this is obtained by computing the heading based on Eq.~\ref{eq:balancefeat} and by passing the result online as the heading objective of the NMPC while keeping the location of the waypoint fixed. To ensure full autonomy of the system a recovery behavior of the system is introduced. If the robot finds itself stuck and unable to move it will send a \textit{RECOVERY} signal to the FSM. First, it tries to solve a situation via common recovery behaviors like in-place rotation and going away from the nearest obstacle (\textit{recovery easy} in the code). Then, if that fails, by leveraging the multi-session mapping capability of the SLAM framework, we start a new mapping session (\textit{recovery hard} in Alg.~\ref{alg:FSM}). The current pose of the robot will be considered the new origin and the FSM will act as previously described. Once a loop closure is found with the previous map, the two will be merged together and rectified.

\begin{small}
\begin{algorithm}
\SetAlgoLined
\KwResult{Exploration}
  state = get NMPC state\;
  \uIf{state = WAITING {\normalfont \textbf{or}} state = GOAL REACHED}{
  \eIf{waypoints {\normalfont \textbf{is not}} empty}{
   \tcp*[h]{\textbf{Second-level activeness}}\\
   optimize next heading\; 
   \tcp*[h]{\textbf{`In' NMPC Third-level activeness}}\\
   send waypoint to NMPC\;
   popfront(waypoints)\;
   }{
   endpoints = require frontier points\;
   \If{endpoints {\normalfont \textbf{is}} empty}{
        \eIf{map graph {\normalfont \textbf{is}} empty}{
        endpoints = 360 rotation in place\;
        }{
        endpoints = random graph node\;
        }
    } 
    \tcp*[h]{\textbf{First-level activeness}}\\
    waypoints = getOptimalSolution(endpoints)\;  
    \tcp*[h]{\textbf{`In' NMPC Third-level activeness}}\\
    send first waypoint to NMPC\;
    popfront(waypoints)\;
  }
 }\uElseIf{state {\normalfont==} RECOVERY}{
    recovery easy\;
    \If{recovery easy fails}{
        recovery hard\;
    }
 }\uElseIf{state {\normalfont==} OPERATING}{
    do nothing\;
 }

\caption{Pseudocode of the finite state machine (FSM) coordinating our proposed active SLAM layers.}
\label{alg:FSM}
\end{algorithm}
\end{small}

	\section{EXPERIMENTS AND RESULTS}
\label{sec:exp}

\begin{figure}[t]
	\centering
	\includegraphics[angle=0,width=0.4\textwidth]{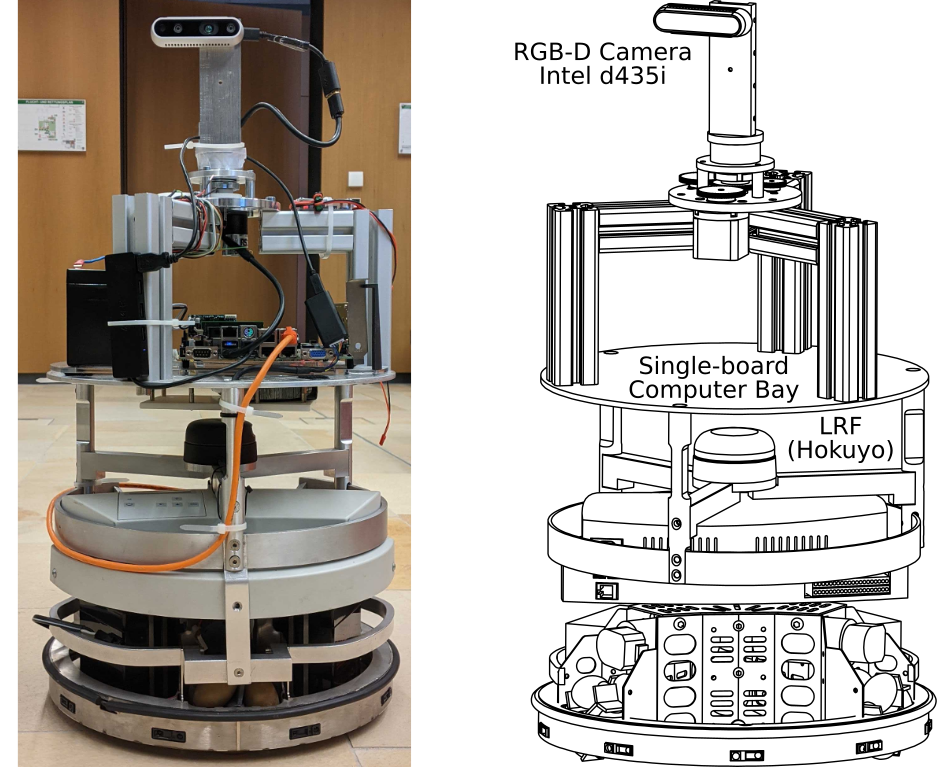}
	\caption{Our robot platform -- Festo Didactic's Robotino with additional structure and hardware. Real robot (left), Gazebo model (right).}
	\label{img:robotschmema}
\end{figure}

Our robot, a Festo Didactic's Robotino 1 (Fig.~\ref{img:robotschmema}), has a 3-wheel omnidirectional drive base. We augmented the robot hardware with an extended physical structure containing a 2D laser range finder (Hokuyo A2M8) an IMU and an RGB-D camera (Intel RealSense D435i) and a single-board computer (Intel® Core™ i7-3612QE CPU @ 2.10GHz). A virtual model of the same robot was created for the simulation experiments.

The choice of an omnidirectional platform is crucial to carry out our proposed continuous camera heading adjustments and the entirety of our method. We recognize the fact that one might consider to use different platforms or hardware like omnidirectional cameras, cameras arrays, or dedicated control mechanism to rotate the camera itself. However, all of these present challenges and introduce errors that would need to be addressed and that might hinder the proposed method. For example, camera arrays suffer from synchronization, calibration, and alignment problems. Omnidirectional/fisheye cameras introduce distortion, are not well suited for indoor environments, and usually lack the presence of a corresponding depth sensor~\cite{7353366}. Finally, dedicated rotating hardware usually lacks the necessary precision and introduces issues related to the relative orientation between the camera, the robot base, and the estimated map and related to the control of the rotation mechanism itself~\cite{9568791}.

\subsection{Simulation Experiments}

\subsubsection{Setup and Implementation}
\label{setupandimplementation}

\begin{figure}[!ht]
    \centering
    \subfloat[AWS's small house]{
    \includegraphics[scale=.22]{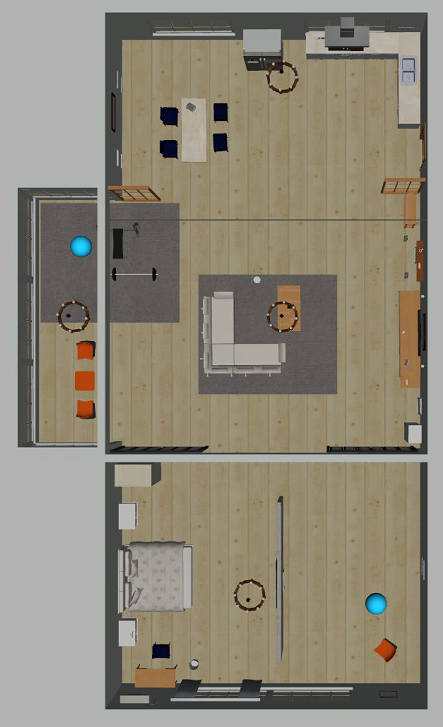}
    \label{fig:aws}
    }
    \quad
    \subfloat[Gazebo's edited Café]{\includegraphics[scale=.28]{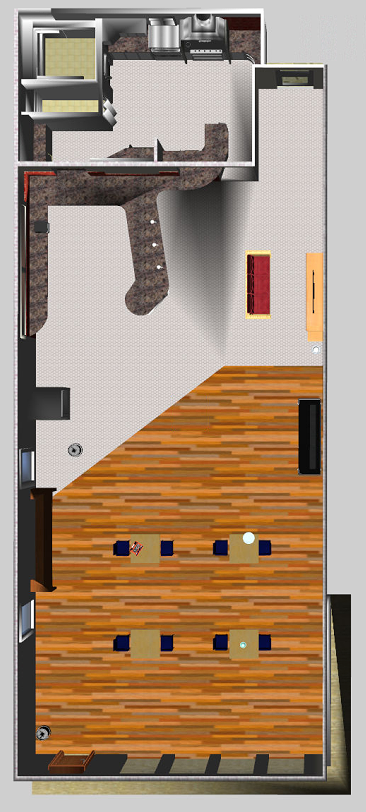}
    \label{fig:gazebo}
    }\caption{Simulated Environments}
    \label{fig:env}
\end{figure}

For validation of the proposed algorithm, we use the Gazebo simulator. We test our method in two different environments: i) AWS Robomaker Small House World\footnote{\url{https://github.com/aws-robotics/aws-robomaker-small-house-world}} (Fig.~\ref{fig:aws}), which is $\sim 180 m^2$, and ii) a modified version of the Gazebo's Café environment (Fig.~\ref{fig:gazebo}) using 3DGems'~\cite{rasouli2017effect}, which is $\sim 200 m^2$.  

We use real-hardware-like parameters for all the sensors in the simulation. The LRF has an angular resolution of $1^{\circ}$ and provides a complete $360^{\circ}$ sweep. The camera is used at a resolution of $848 \text{x} 480$. Its horizontal FOV is $69.4^{\circ}$, and its maximum sensing depth is $4$ meters. For all experiments, the maximum translation speed of the robot is set to $1$ m/s, while the maximum angular speed is $1$ rad/s. The ground truth maps were obtained by using \textit{pgm\_map\_creator}\footnote{\url{https://github.com/hyfan1116/pgm_map_creator}} and edited to remove unwanted artefacts. The current state estimate of the robot is obtained through a classical sensor fusion technique that uses the IMU, the odometry of the robot, and the ICP odometry obtained through the LRF. The loop closures events are used as a low-frequency `GPS'-like sensor. We use the widely adopted ACADO framework~\cite{Houska2011a} to setup the optimization and used qpOASES as a solver. The horizon is made of $20$ steps of $0.1$ seconds each. We consider ten obstacles in our implementation and their position is treated as online data in the framework. The obstacles are detected through the \textit{costmap\_converter}\footnote{\url{https://github.com/rst-tu-dortmund/costmap_converter}} package which provides us the estimated obstacle's velocity and convex hull. The nearest vertex is used for each one of the detected obstacles. Those are then sorted based on their status (dynamic or static) and their distance to the robot. Priority is given first to the dynamic ones, and then to the nearest static ones, with respect to the state predicted at each time step by the previous iteration of the NMPC. Our initial state is the most recent estimate of the robot pose provided by the odometry estimation system. Maps have a cell size of $0.05$ m and are compared w.r.t. the ground truth using the balanced accuracy score (BAC) on the three classes free, occupied and unknown. The comparison is also made based on the root mean squared absolute trajectory error (ATE RMSE) and the number of loop closures triggered per meter of distance travelled. Map entropy is monitored throughout the entire experiment and presented as normalized entropy w.r.t.\ the actual explored area. Wheels' total rotation per meter travelled is also used to further show the energy-saving benefits of our method by showing the actual movement taken by them throughout the experiment. Finally, the area exploration amount per meter travelled by the robot is employed to compare energy efficiency between the algorithms and capture the overall movement that the robot takes to explore a comparable amount of area.

We use RTAB-Map~\cite{labbe2019rtab} as our V-SLAM back-end, which is a graph-feature-based visual SLAM framework. The 2D occupancy map is generated by the 2D projection of the 3D octomap built through our RGB-D sensor and used for navigation and evaluation. This allows us to map the obstacles that would be otherwise hidden to the 2D LRF.

\subsubsection{Methods naming and description}

As our overall active SLAM approach is extremely modular (different possible ways to compute path utility and the three levels of activeness), we combine our contributions to develop various comparisons. 

First, we consider 2 additional path's utility computation methods. These are i) OL: using only the optimized goal position instead of the waypoints on the path and ii) INTER: the orientation of the robot is simply interpolated along the trajectory to the goal position and the total utility is computed as a plain sum of the utilities. Both of these neglect our \textit{first-level activeness}. Active SLAM approaches using these for utility computation are prefixed by OL and INTER, respectively. Approaches using our complete procedure of global path generation in (Subsec.~\ref{subsec:globalpath}) are prefixed with A.
\
Based on the combination of activeness levels and path utility computation approach, we performed experiments with the following methods -- A (our complete approach), A\_1, OL\_0, OL\_1, OL\_1\_3, OL\_2, OL\_2\_3, and INTER\_0. The numbers in the suffix signify the activeness levels enabled, with 0 referring to none of them enabled. For simplicity, if no number is specified all active levels are enabled. We performed a longer time duration version of our complete approach named A\_L and a classic weighted sum approach named A\_S. Finally, we also provide experiments with 2 additional methods, where our complete approach is used, but instead of using the plain Shannon entropy scheme ($u_1$), we use $u_2$ and $u_3$, as introduced in Subsec.~\ref{subsec:utility}. These methods are named A\_O ($u_2$) and A\_DW\_O ($u_3$). Note that OL\_0 is similar to~\cite{Dai2020}, since it chooses the long-term goal based on the heading optimization for the last waypoint. Waypoints' headings are not being optimized since, as explained also in~\cite{Dai2020}, their method generates a very low number of intermediate steps in contrast to our 1-m apart approach. INTER\_0 represents a non-optimized approach toward the frontier, similarly to~\cite{CarrilloDamesKumarCastellanos2017}. As stated above, we use a plain Shannon's entropy formulation, differently from what was proposed in~\cite{CarrilloDamesKumarCastellanos2017}. Moreover, here it is the RGB-D sensor that performs the mapping operation, and not a 2D laser sensor. Note that the information gathered along the interpolated trajectory does not modify in any way the path or the planned robot headings, making this an ideal comparison as it represents a greedy approach. Those are our considered baselines.

Simulation results are the average over 20 different successful trials of ten minutes each for each method  with the same starting location on the map. Variability in trials is achieved as a result of different trajectories the robot takes in each one of the runs. Since RTABMap's update rate is not fixed but varies based on the movement of the robot, the plots are presented after a bucketing procedure in time windows of 2 seconds, eventually filled with previous values. 

\subsubsection{Results}

\begin{figure*}[!ht]
    \centering
    \subfloat[Path length evolution]{
    \includegraphics[width=.8\columnwidth]{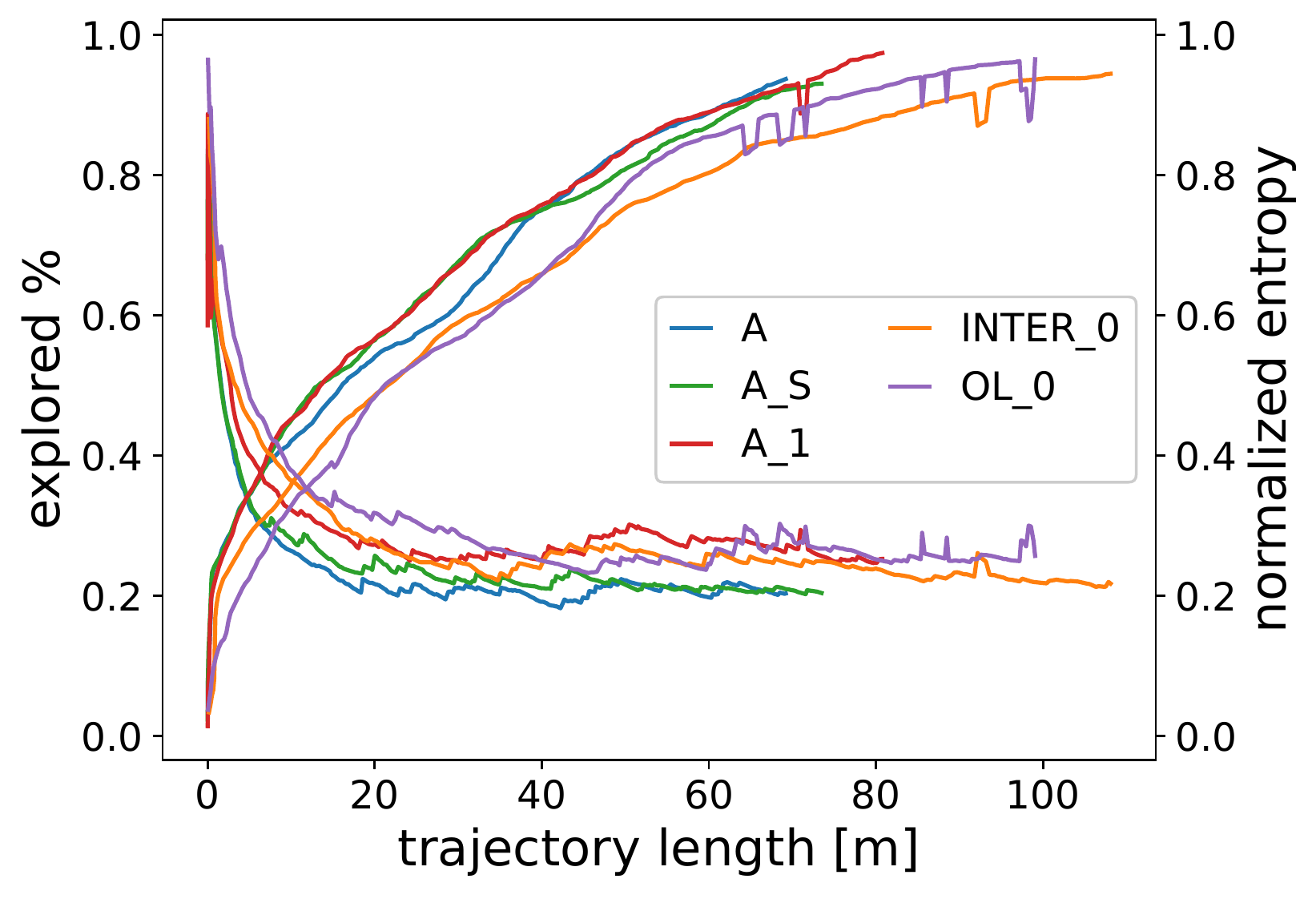}
    }
    \subfloat[Time evolution]{\includegraphics[width=.8\columnwidth]{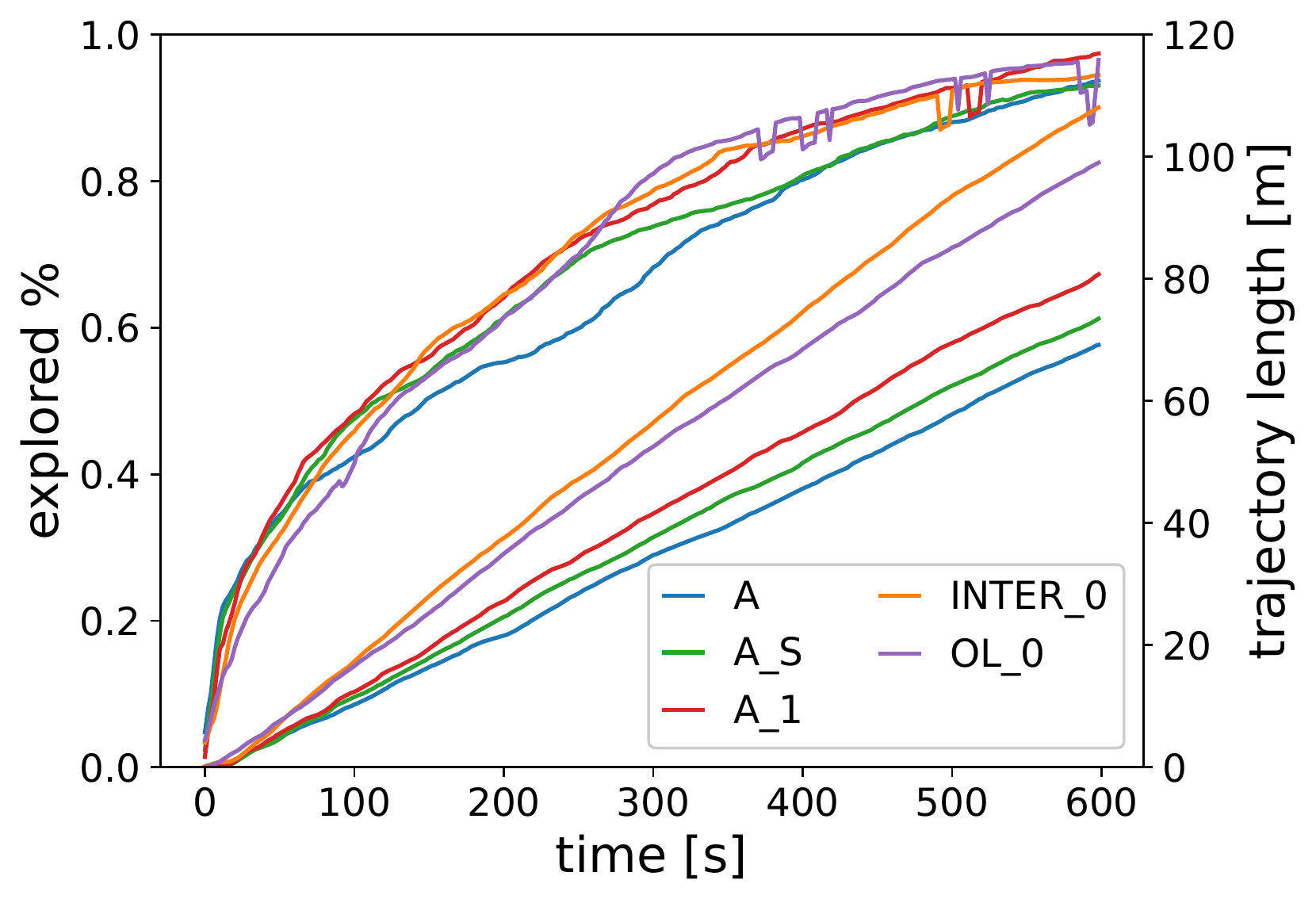}
    }\caption{Café environment: comparison of proposed methods.}
    \label{fig:gazesoa}
\end{figure*}

\begin{figure*}[!ht]
    \centering
    \subfloat[Path length evolution]{
    \includegraphics[width=.8\columnwidth]{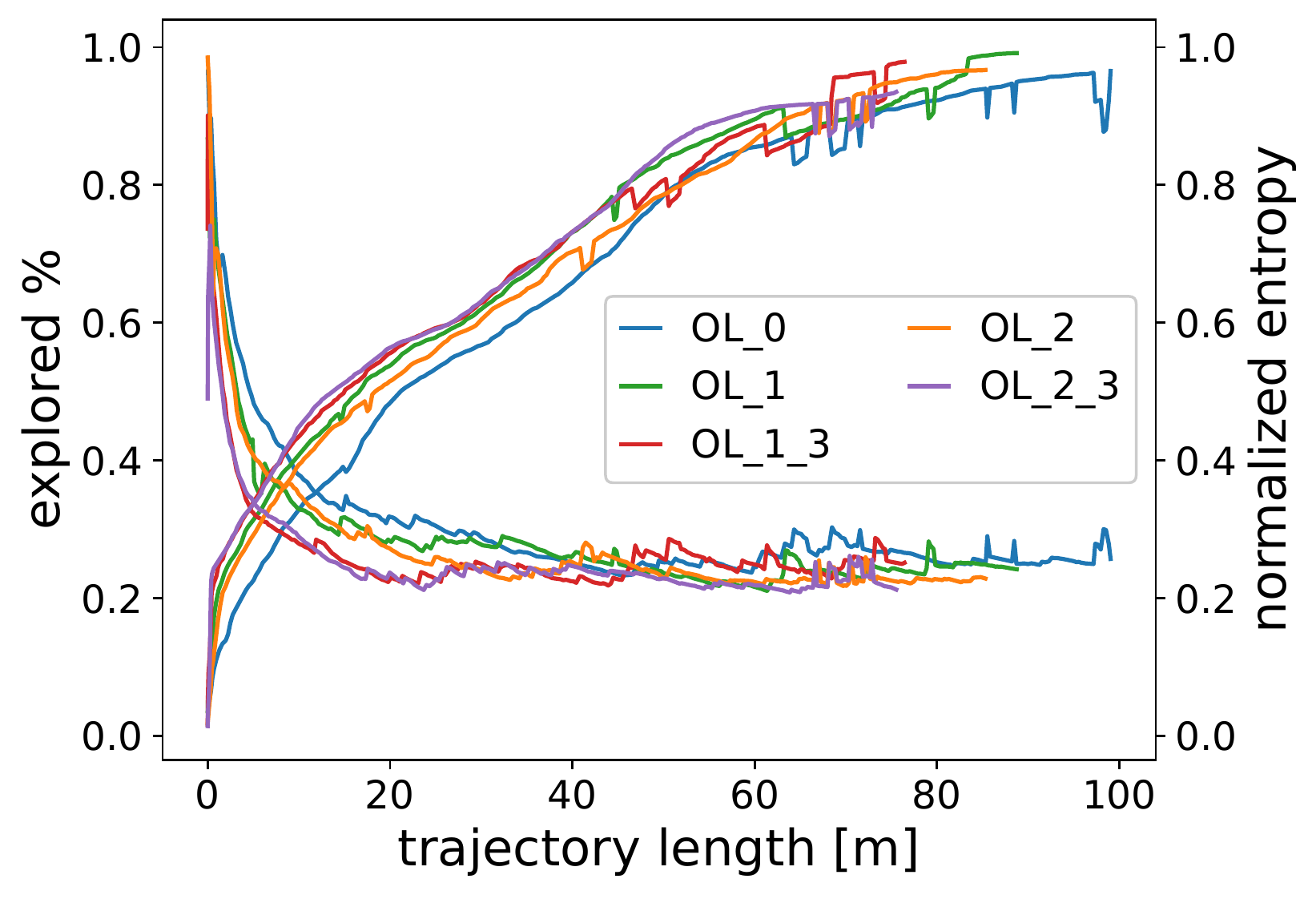}
    }
    \subfloat[Time evolution]{\includegraphics[width=.8\columnwidth]{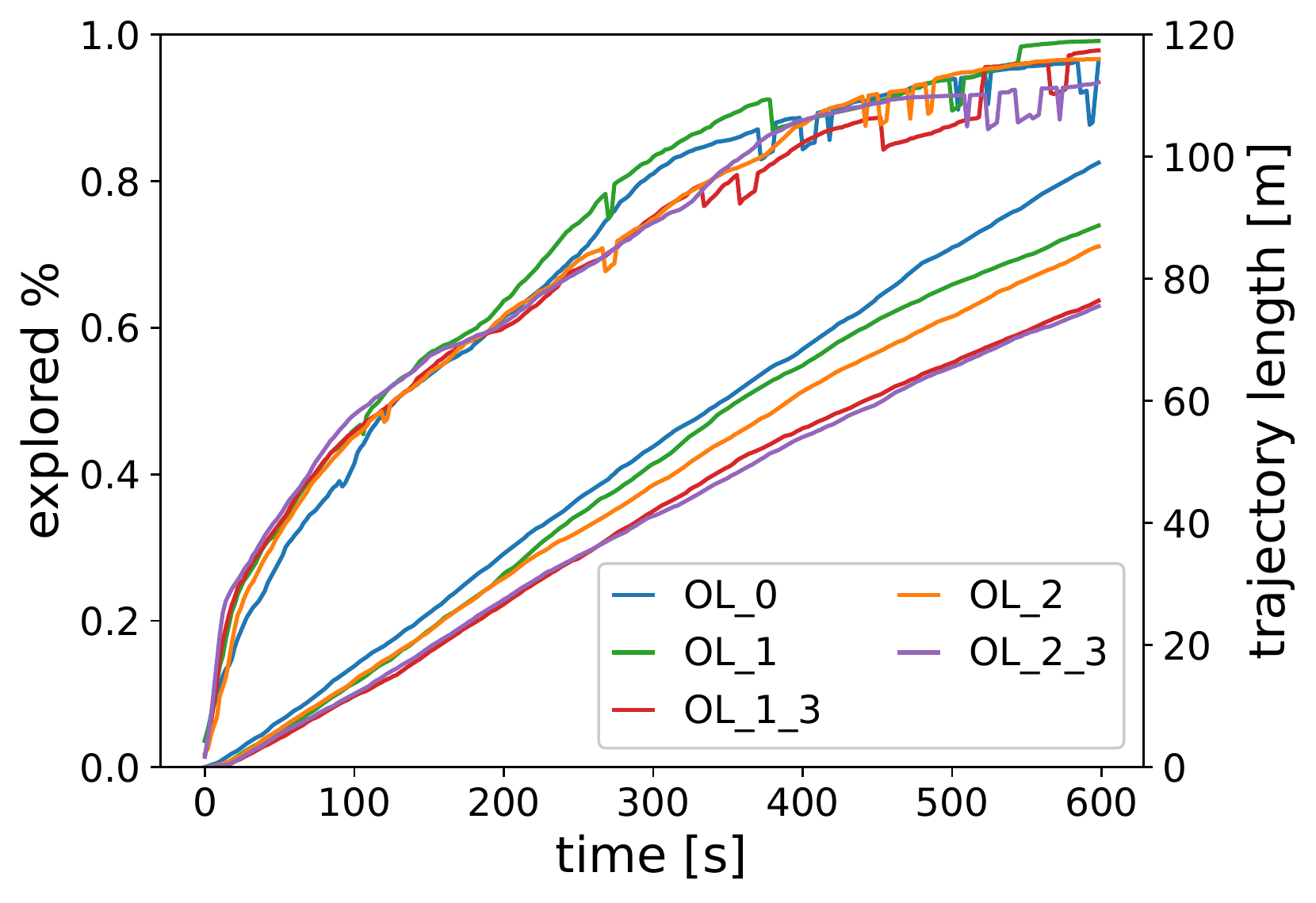}
    }\caption{Café environment: comparison of OL-type methods}
    \label{fig:gazeolact}
\end{figure*}

\begin{figure*}[!ht]
    \centering
    \subfloat[Path length evolution]{
    \includegraphics[width=.8\columnwidth]{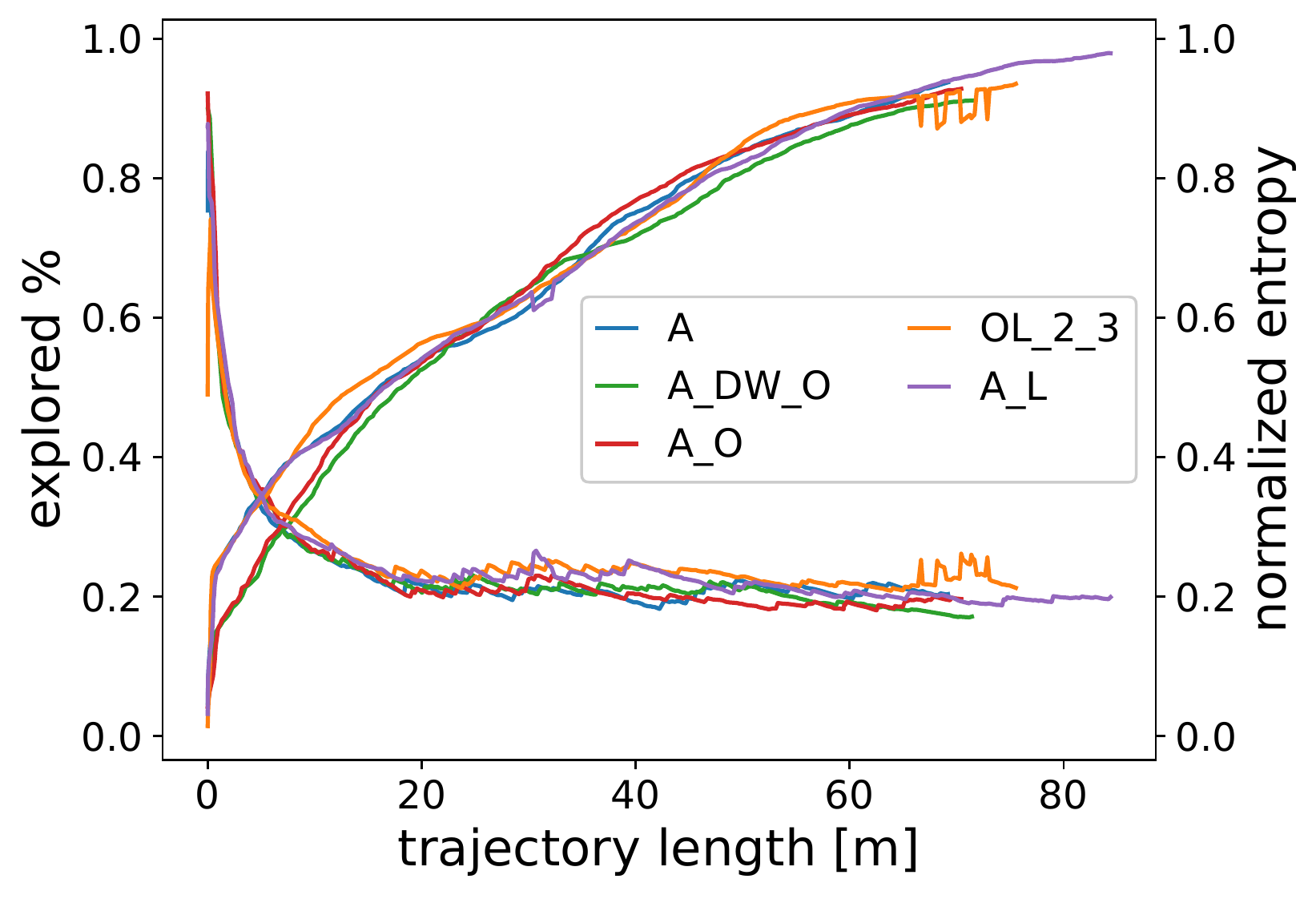}}
    \subfloat[Time evolution]{\includegraphics[width=.8\columnwidth]{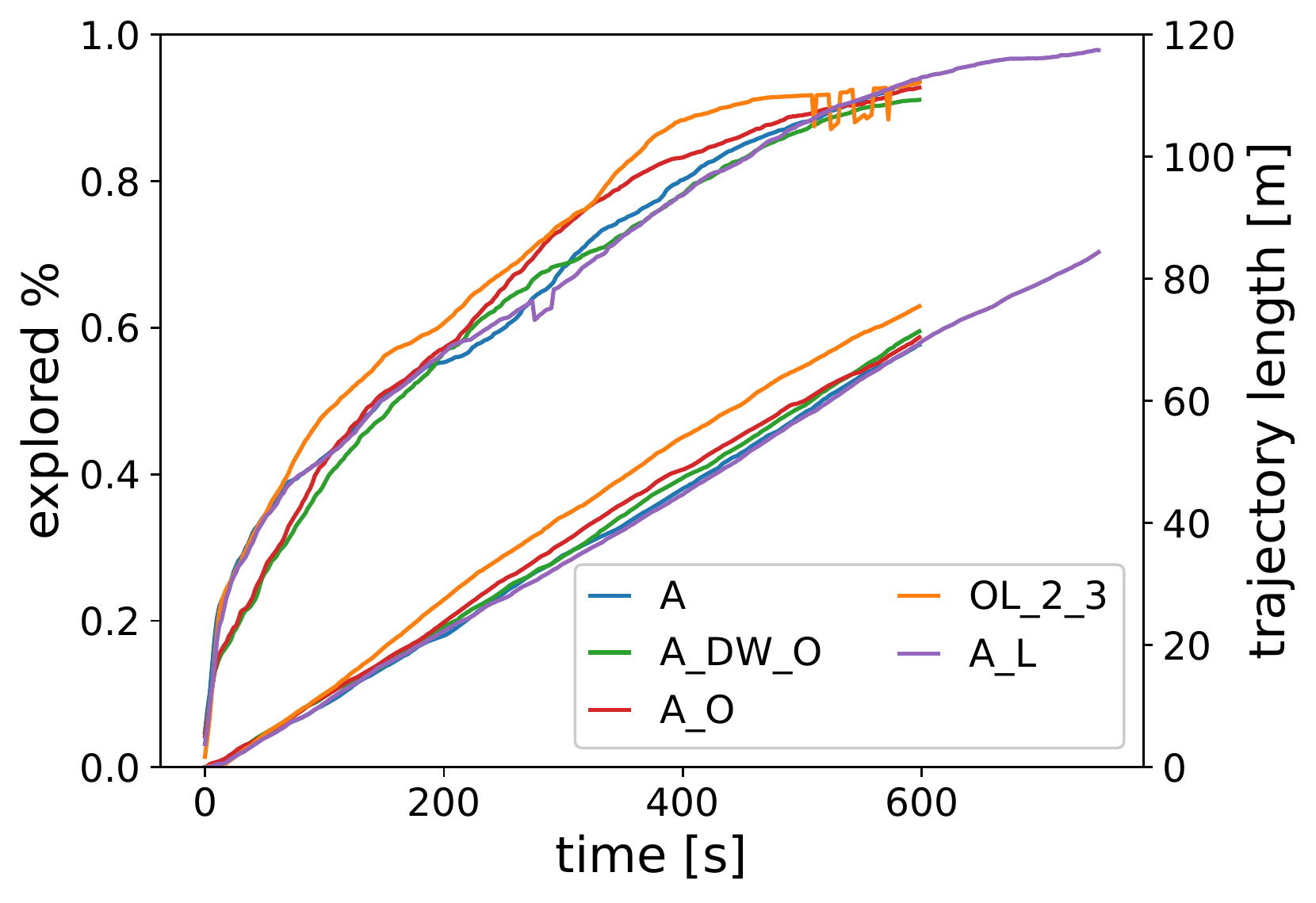}}
    \caption{Café environment: comparison of different utilities}
    \label{fig:gazeours}
\end{figure*}

\subsubsection*{A. Baselines comparison}
\begin{figure*}[!ht]
    \centering
    \subfloat[Path length evolution]{
    \includegraphics[width=.8\columnwidth]{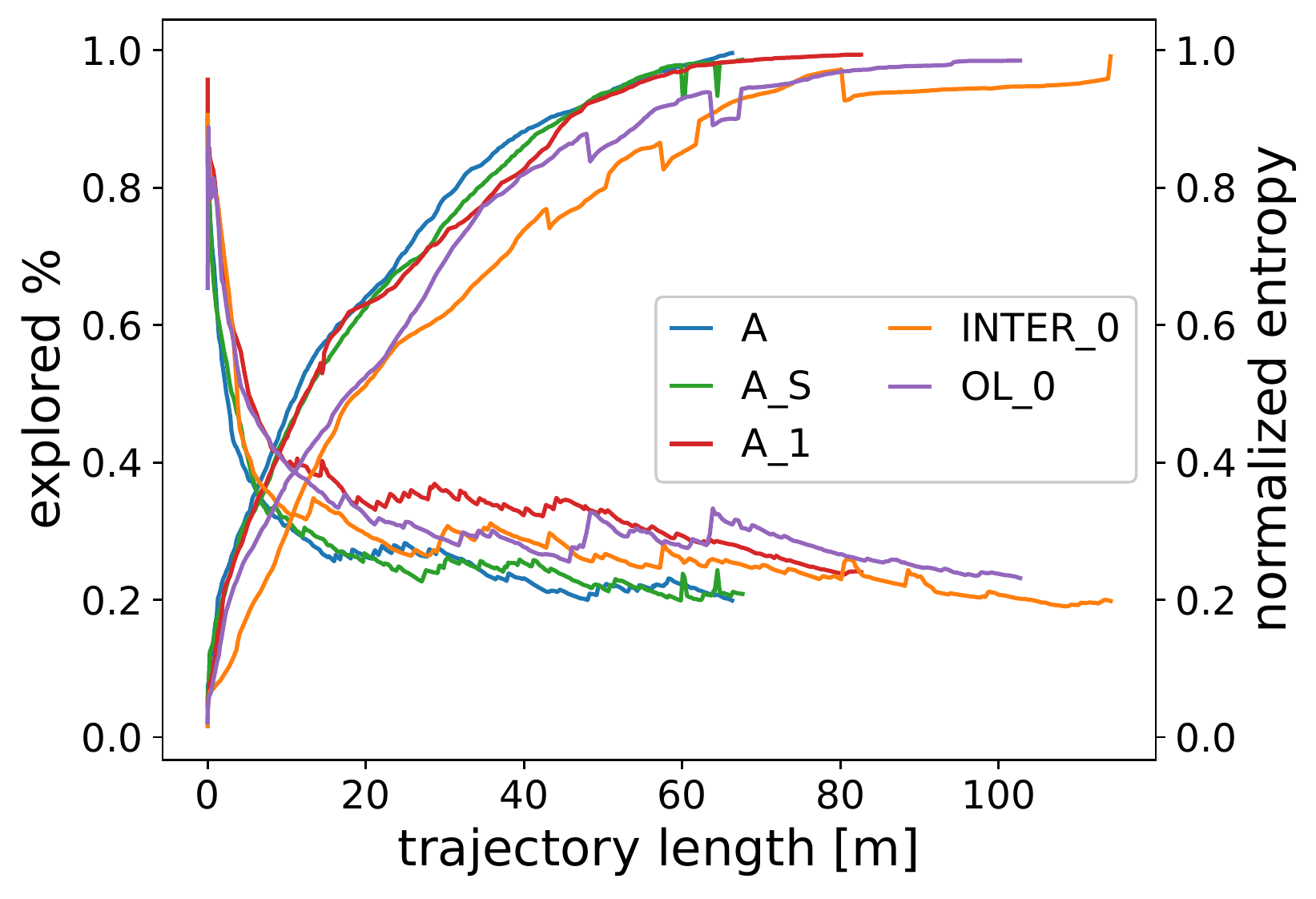}
    }
    \subfloat[Time evolution]{\includegraphics[width=.8\columnwidth]{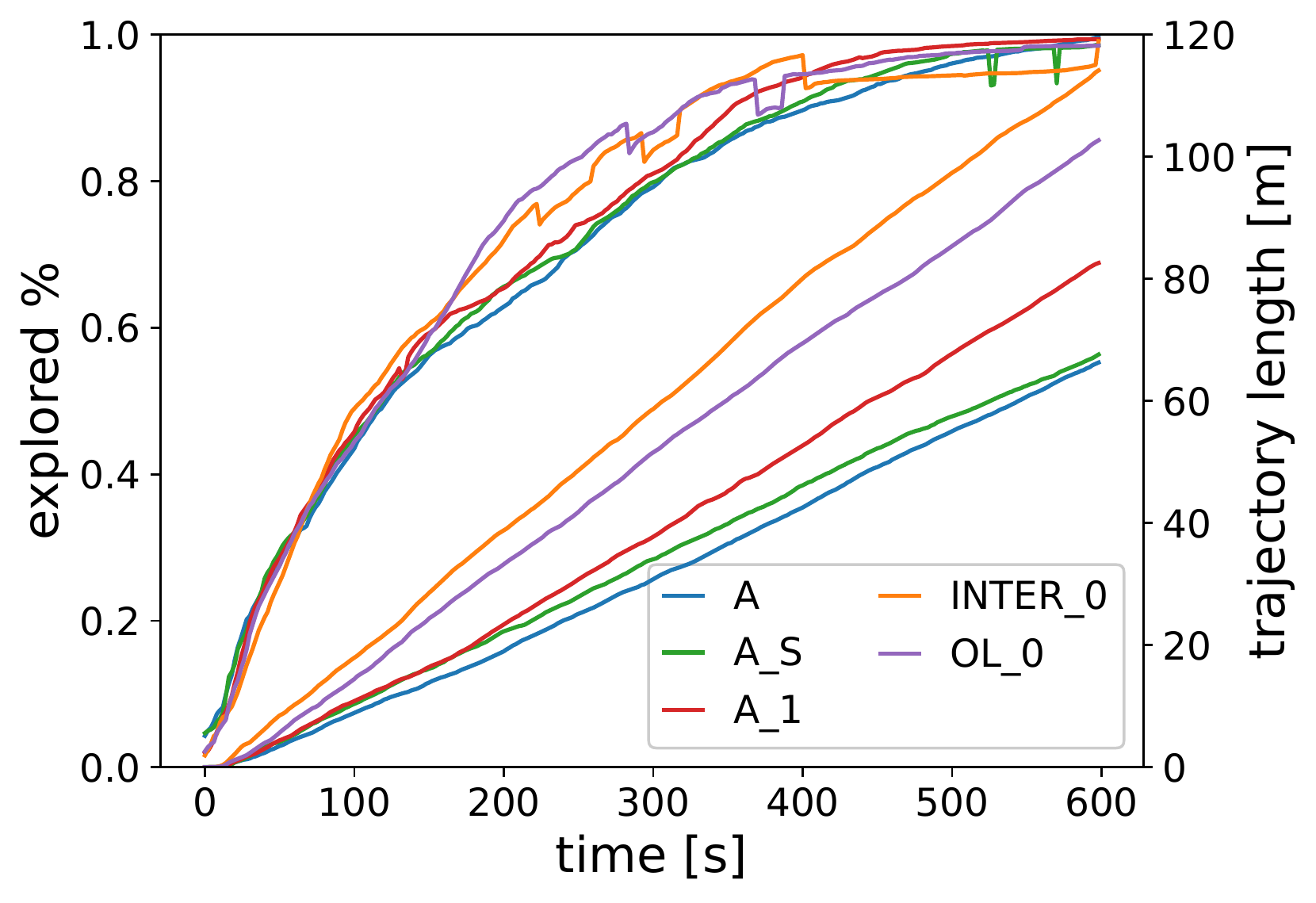}
    }\caption{Small house: comparison of proposed methods.}
    \label{fig:awssoa}
\end{figure*}

We first compare our full method A with A\_1 (i.e. pre-plan waypoints' BV), OL\_0, and INTER\_0 (baselines). Results are equivalent for both environments as shown by~\tab{tab:gazeours},~\tab{tab:awsours},~\fig{fig:gazesoa}, and~\fig{fig:awssoa}. Notable differences can be observed in the trajectory length. Our active approaches (A and A\_1) successfully explore the environment with much shorter trajectories. Indeed, OL\_0 ones are $\sim33\%$ and $\sim19\%$ longer while for INTER\_0 we have $\sim39\%$ and $\sim24\%$ both w.r.t. A and A\_1. Even considering the wheels' rotation per meter travelled by the robot, which is higher for the active methods (the highest difference is $\sim11\%$ between A and INTER\_0) the advantage is clear. A higher amount of wheels' overall movement for the active methods is expected, since we enforce an (almost) continuous rotational movement with our approach.

As for absolute trajectory error, we can notice that A, A\_1, and OL\_0 show comparable results. However, INTER\_0, thanks to the smooth movement of the robot associated with the interpolated trajectory, exhibits a lower ATE RMSE. Nonetheless, this does not translate into any clear benefit over the final entropy, exploration amount, or exploration speed compared to all other algorithms. 

Looking now at the normalized entropy, despite completing the exploration earlier, thus having time to refine the result, INTER\_0 and OL\_0 perform similarly in terms of BAC w.r.t. our active algorithms A and A\_1. However, both show higher normalized entropy in comparison to A showing how our approach better maps the environment on the first pass. Moreover, INTER\_0 in general shows the lowest exploration profile and, despite the longest trajectory length, is not reaching a full exploration.

Regarding loop closures, we can see how A generates more of them, compared to INTER\_0 and OL\_0, despite a much lower path length. On the other hand, A\_1 shows worse results in this metric. However, one must notice that A\_1, in contrast to INTER\_0 and OL\_0, does not have the possibility to navigate in an almost-fully explored environment toward the end of the experiment. This could be one of the reasons behind this result alongside the continuous rotational movement. Anyway, given the re-observation possibilities and the movement smoothness of both INTER\_0 and OL\_0 one would expect them to have higher loop closures. The fact that this is not the case further proves the benefits brought by our approach. Moreover, we will show in the ablation studies how the third level of activeness especially influences this metric, which will justify the difference of A\_1 w.r.t. A.

Finally, in~\fig{fig:gazesoa} and~\fig{fig:awssoa}, we can notice how, despite having a much higher exploration area per meter travelled and a continuous rotational movement, our active method A continuously keeps a much lower normalized entropy. By design, the second and (especially) the third level of activeness cause the robot to make more rotational movements. Such movements are known to increase the uncertainty in pose estimates of the robot. However, as is evident, in our active approach they still do not adversely affect the robot localization. 

\begin{table}[!hb]
\centering
\setlength\tabcolsep{3pt}
\resizebox{\columnwidth}{!}{\begin{tabular}{lcccccccc}
\hline
{}        & \multicolumn{2}{c}{BAC}                            & \multicolumn{2}{c}{ATE RMSE}                                    & \multicolumn{2}{c}{\begin{tabular}{@{}c@{}}Per  meter \\ loops\end{tabular}} & \multicolumn{2}{c}{\begin{tabular}{@{}c@{}}Per  meter \\ wheels' rotation\end{tabular}}                     \\ \hline
          & \multicolumn{1}{c}{avg} & \multicolumn{1}{c}{std} & \multicolumn{1}{c}{avg [m]} & \multicolumn{1}{c}{std [m]} & \multicolumn{1}{c}{avg} & \multicolumn{1}{c}{std}
          & \multicolumn{1}{c}{avg [r/m]} & \multicolumn{1}{c}{std [r/m]} \\\hline 
A      &  $0.771$   & $0.022$ &  $0.059$    & $0.028$ &  $4.965$  &$0.596$ & $65.781$ & $4.532$ \\
A\_L & $0.789$ & $0.019$ & $0.047$ & $0.023$ & $5.149$ & $0.597$ & $68.116$ & $4.944$ \\
A\_S & $0.767$ & $0.021$ & $0.064$ & $0.062$ & $5.146$ & $0.491$ & $66.285$ & $5.829$ \\
INTER\_0  &  $0.772$   & $0.024$ &  $0.038$    & $0.022$ &  $4.303$  & $0.393$ & $59.134$ & $2.672$ \\
OL\_0     &  $0.780$   & $0.030$ &  $0.060$    & $0.058$ &  $4.198$  &$0.580$ & $60.615$ & $4.942$ \\
A\_1 &  $0.776$   & $0.026$ &  $0.063$    & $0.031$ &  $3.211$  & $0.440$ & $63.705$ & $6.018$ \\
OL\_1    & $0.802$ & $0.018$ & $0.051$ & $0.025$ & $3.838$ & $0.640$ & $61.755$ & $4.546$\\
OL\_1\_3 & $0.792$ & $0.028$ & $0.078$ & $0.046$ & $4.986$ & $0.469$ & $69.080$ & $5.928$\\
OL\_2      & $0.785$ & $0.031$ & $0.084$ & $0.054$ & $3.984$ & $0.867$ & $67.288$ & $13.683$ \\
OL\_2\_3   & $0.776$ & $0.027$ & $0.066$ & $0.057$ & $5.120$ & $0.650$ & $68.506$& $5.868$\\
A\_O     & $0.772$ & $0.030$ & $0.041$ & $0.011$ & $5.302$ & $0.587$ & $66.313$ & $8.356$\\
A\_DW\_O & $0.765$ & $0.022$ & $0.046$ & $0.024$ & $5.144$ & $0.514$ & $64.594$ & $4.731$\\
\end{tabular}}
\caption{Café environment results}
\label{tab:gazeours}
\end{table}

\subsubsection*{B. Using weighted average}
Now we want to compare the effect that the weighted average has on our full approach. To do so, we consider A that uses the proposed weighted average scheme, and A\_S which applies the classic weighted sum. In the Café world performances are quite similar, with a slightly lower ATE for the weighted average approach as can be observed in~\tab{tab:gazeours}. Instead, when tested in the AWS environment using the weighted average scheme, there are significant gains throughout all the metrics, except for the loop closures' one (see ~\tab{tab:awsours}). The two approaches in both the considered worlds have similar exploration profiles and trajectory lengths (see~\fig{fig:gazesoa} and~\fig{fig:awssoa}). However, the normalized entropy resulting from our approach is lower during the majority of the duration of the experiment. This result could be attributed to the ability of the weighted average to better capture the utility of the whole movement that will bring the robot to the frontier point, rather than favouring single high-utility spikes.

\begin{table}[!ht]
\centering
\setlength\tabcolsep{3pt}
\resizebox{\columnwidth}{!}{\begin{tabular}{lcccccccc}
\hline
{}        & \multicolumn{2}{c}{BAC}                            & \multicolumn{2}{c}{ATE RMSE}                                    & \multicolumn{2}{c}{\begin{tabular}{@{}c@{}}Per  meter \\ loops\end{tabular}} & \multicolumn{2}{c}{\begin{tabular}{@{}c@{}}Per  meter \\ wheels' rotation\end{tabular}}                     \\ \hline
          & \multicolumn{1}{c}{avg} & \multicolumn{1}{c}{std} & \multicolumn{1}{c}{avg [m]} & \multicolumn{1}{c}{std [m]} & \multicolumn{1}{c}{avg} & \multicolumn{1}{c}{std}
          & \multicolumn{1}{c}{avg [r/m]} & \multicolumn{1}{c}{std [r/m]} \\\hline 
A      & $0.818$ & $0.015$ & $0.051$ & $0.020$ & $4.692$ & $0.382$ & $63.043$ & $2.540$ \\
A\_L    & $0.815$ & $0.024$ & $0.052$ & $0.021$ & $4.880$ & $0.704$ & $66.649$ & $9.372$\\
A\_S & $0.788$ & $0.026$ & $0.101$ & $0.077$ & $4.722$ & $0.526$ & $67.236$ & $9.299$ \\
A\_1   & $0.805$ & $0.023$ & $0.060$ & $0.031$ & $2.923$ & $0.477$ & $59.382$ & $2.851$ \\
INTER\_0 & $0.823$ & $0.011$ & $0.034$ & $0.013$ & $3.791$ & $0.386$ & $56.721$ & $2.886$\\
OL\_0      & $0.808$ & $0.016$ & $0.063$ & $0.045$ & $4.057$ & $0.658$ & $61.687$ & $20.745$\\
OL\_1   & $0.816$ & $0.029$ & $0.052$ & $0.037$ & $3.402$ & $0.525$ & $62.389$ & $9.463$ \\
OL\_1\_3 & $0.815$ & $0.021$ & $0.058$ & $0.023$ & $4.665$ & $0.426$ & $68.163$ & $19.557$\\
OL\_2     & $0.820$ & $0.019$ & $0.079$ & $0.062$ & $3.495$ & $0.470$ & $66.980$ & $12.378$\\
OL\_2\_3   & $0.818$ & $0.016$ & $0.067$ & $0.045$ & $4.849$ & $0.642$ & $65.048$ & $9.937$\\
A\_O    & $0.821$ & $0.015$ & $0.048$ & $0.022$ & $4.791$ & $0.533$ & $63.679$ & $3.588$\\
A\_DW\_O & $0.806$ & $0.024$ & $0.047$ & $0.022$ & $4.738$ & $0.569$ & $63.564$ & $3.679$\\
\end{tabular}}
\caption{Small house results}
\label{tab:awsours}
\end{table}

\subsubsection*{C. Ablation studies}

Since, as shown previously, OL\_0 outperforms INTER\_0, we check how our active components influence this method by comparing OL\_1, OL\_1\_3, OL\_2, and OL\_2\_3. In this, we apply each active component (\textbf{1}st, \textbf{2}nd, and \textbf{3}rd level of activeness) separately, leveraging the modularity of our algorithm, on top of OL\_0. 

\begin{figure*}[!ht]
    \centering
    \subfloat[Path length evolution]{
    \includegraphics[width=.8\columnwidth]{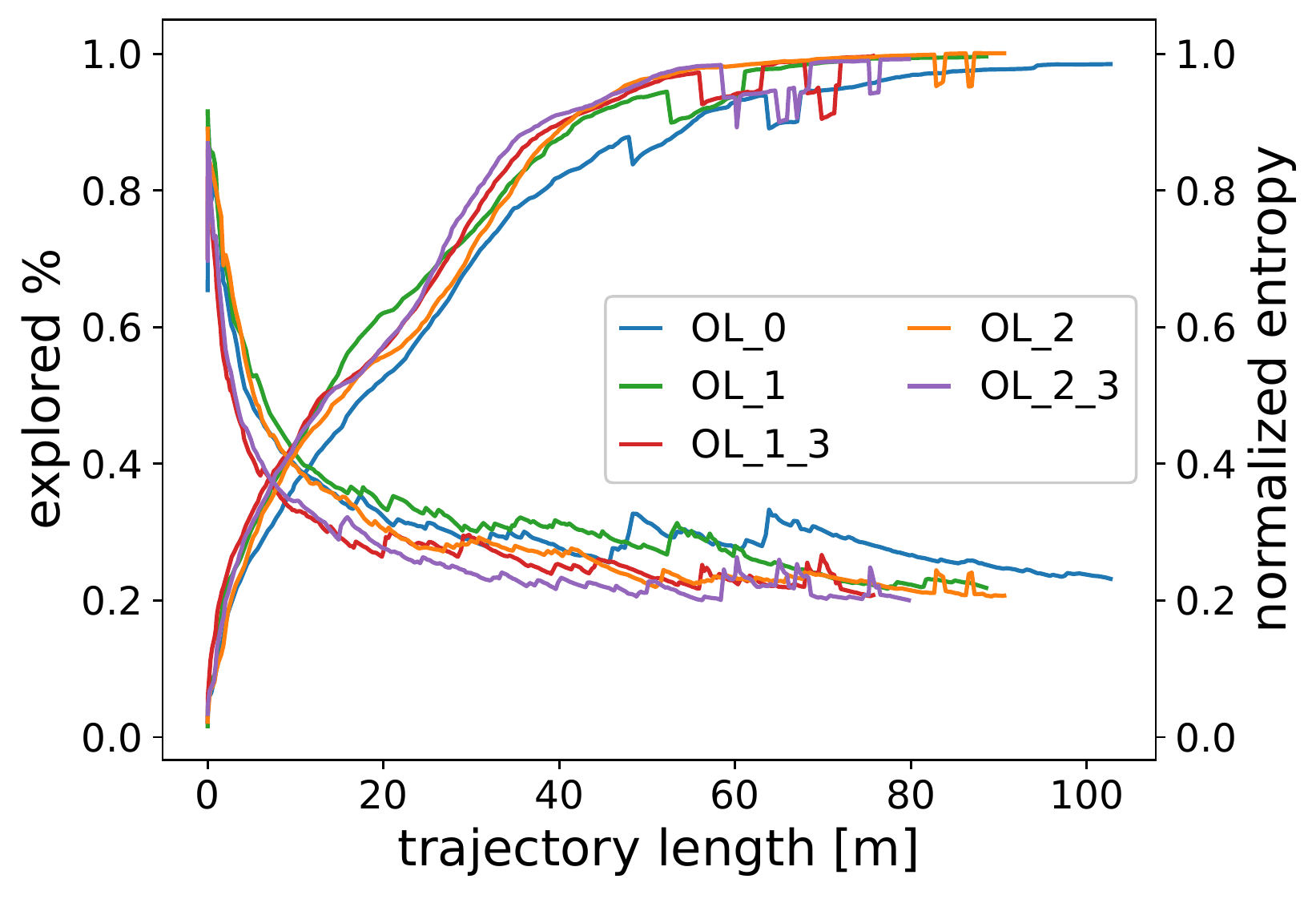}
    }
    \subfloat[Time evolution]{
    \includegraphics[width=.8\columnwidth]{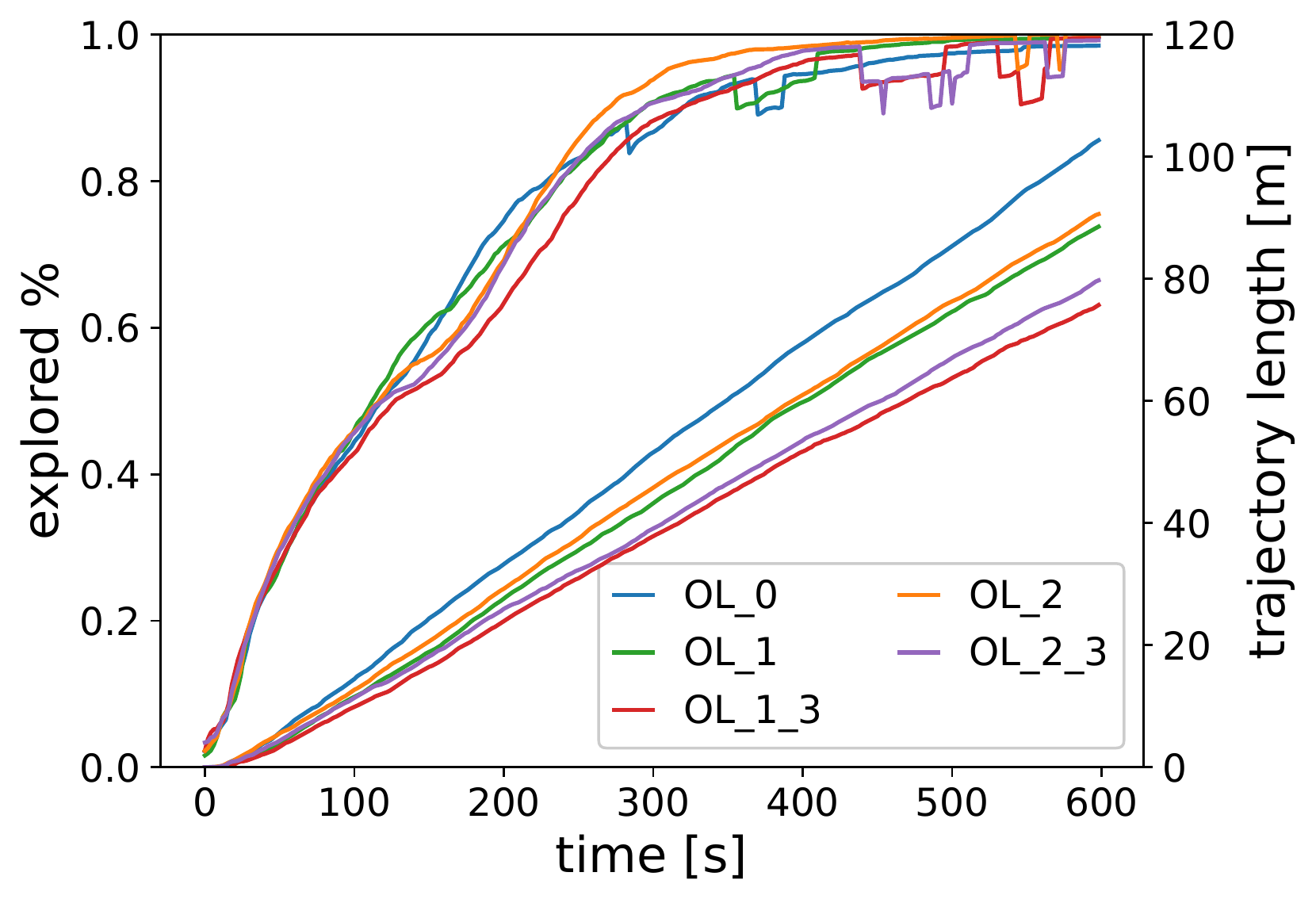}
    }\caption{Small house: comparison of OL-type methods}
    \label{fig:awsolact}
\end{figure*}

We see from Fig.~\ref{fig:gazeolact} and~\fig{fig:awsolact} that all methods outperform OL\_0. It is clear that path lengths are majorly affected when the third level of activeness is enabled with a steep reduction within the same time window. Tracking features causes a slight delay in reaching the desired orientation at every waypoint. This could be solved with more careful tuning of the weights of the third level of activeness or of the NMPC. 

As expected, pre-fixing the orientation along the selected path (OL\_1) or doing so just for the subsequent waypoint (OL\_2) has little influence over the final trajectory length since the difference in the computational effort is minimum. 

Overall, OL\_1\_3 and OL\_2\_3 show better performance in the majority of the metric comparisons w.r.t. OL\_1 and OL\_2. Notably, we can see how, as expected, the loop closure amounts increase when the third level of activeness is enabled. 
Moreover, since the two are designed to perform similar actions, OL\_1\_3 and OL\_2\_3 exhibit similar performances. However, it seems that OL\_2\_3 has a slight advantage by looking at BAC, ATE, and loop closure amounts.

\begin{figure*}[!ht]
    \centering
    \subfloat[Path length evolution]{
    \includegraphics[width=.8\columnwidth]{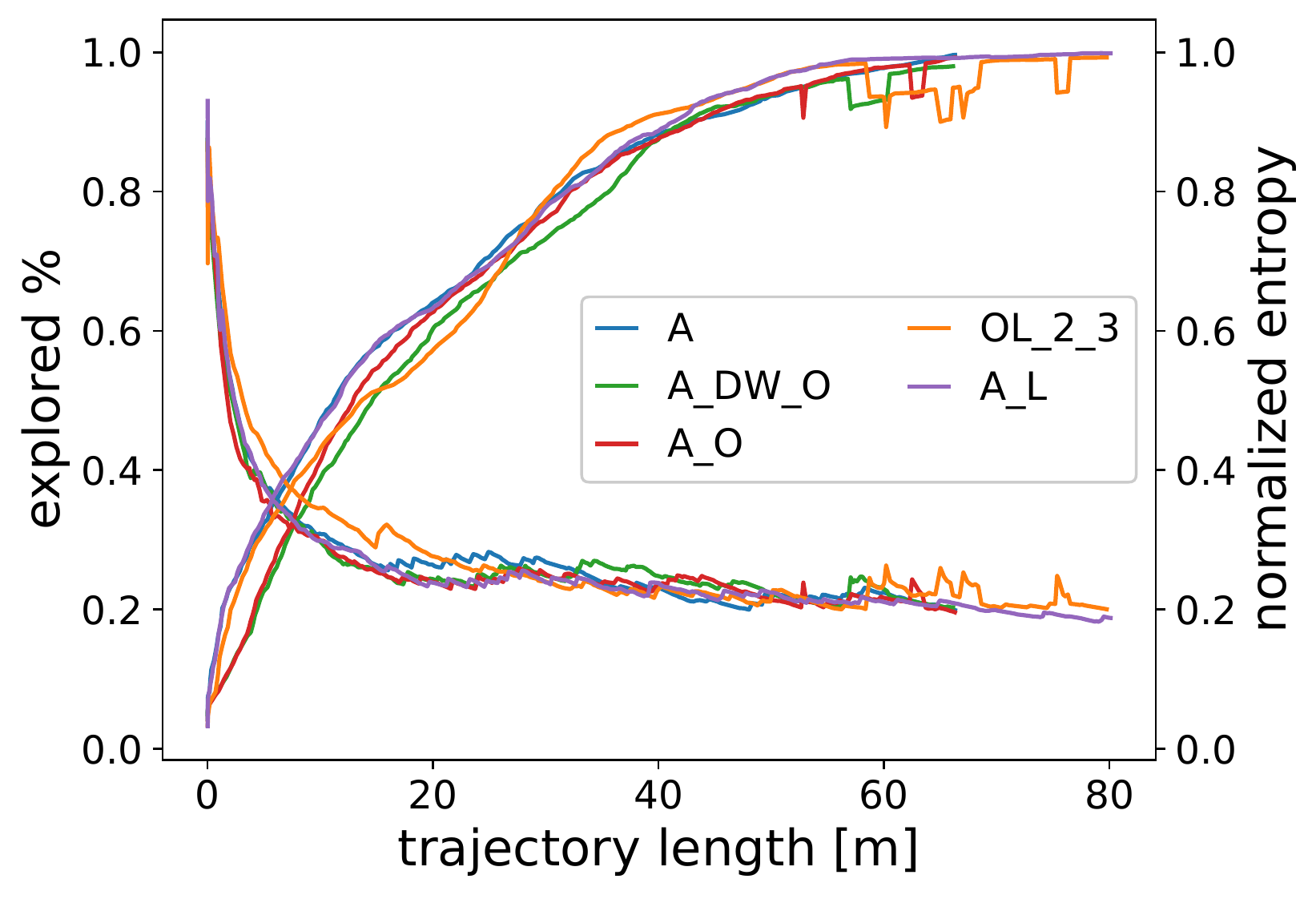}
    }
    \subfloat[Time evolution]{
    \includegraphics[width=.8\columnwidth]{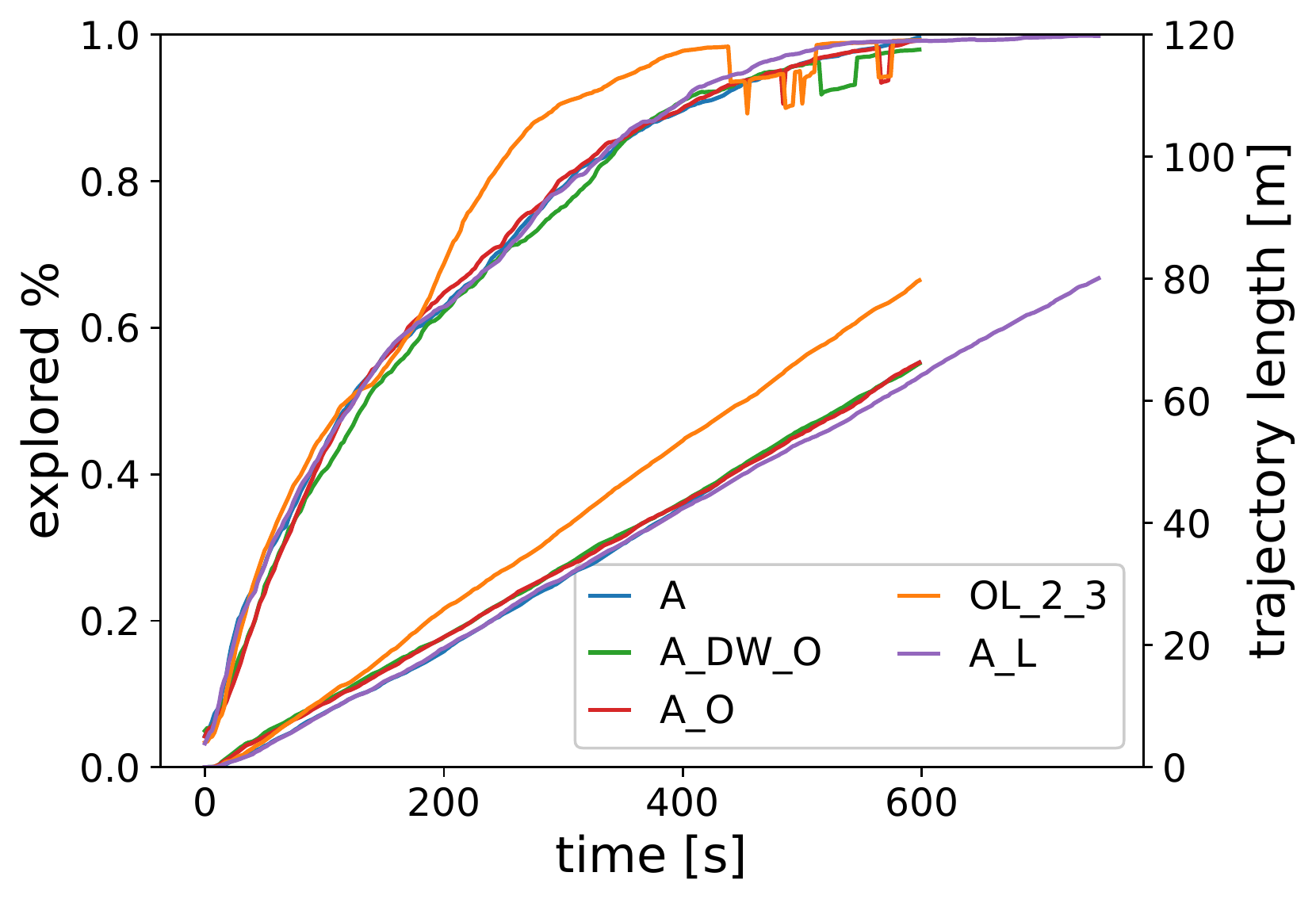}
    }
    \caption{Small house: comparison of different utilities}
    \label{fig:awsours}
\end{figure*}

Comparing OL\_*\_3 with OL\_0 further proves that the continuous rotational movement introduced through the active levels does not harm the map quality or the trajectory reconstruction error. 

\subsubsection*{D. Utilities and activeness comparison}
Finally, in our last comparison, shown in~\fig{fig:gazeours} and~\fig{fig:awsours}, we compare OL\_2\_3 against our full approach A, the proposed utilities (A\_O and A\_DW\_0), and A\_L.  Recall that A\_L is an extended ($12.5$ minutes) version of our full approach A.
We observe that all our approaches, when compared to OL\_2\_3, can explore the environment \textit{but} with shorter paths ($\approx 5-12\%$). Moreover, even considering the wheels' rotation, we notice similar or better performance. This excludes a higher energy consumption due to the continuous rotation.

Our full approaches keep the normalized entropy lower both \textit{during} and at the \textit{end} of the experiments. The exploration speed and amount are overall comparable but if we look at the time evolution, OL\_2\_3 seems to flatten earlier than expected. The ATE RMSE is, in general, better for A(\_L), A\_O, and A\_DW\_0, as compared to OL\_2\_3. Without loss of generalization, similar considerations can be made w.r.t. OL\_1\_3 given the similarities pointed out before. Notice also how all the OL methods are more prone to failures and to the necessity of recovery, further showing the necessity of considering the whole path.

We can observe that both methods A\_O and A\_DW\_0, as expected, are linked to lower normalized entropy and lower ATE. The presence of obstacles in the utility function seems to improve the overall system. This justifies the introduction of such a term in the utility function for an active V-SLAM approach. Since the obstacle cells are the ones harder to refine, and for which the system must be more vigilant, exploiting them directly in the utility computation is beneficial. Using the second formulation of the utility function ($u_2$, A\_O), performs marginally better on both BAC and ATE against the third formulation ($u_3$, A\_DW\_O). However, A\_DW\_O shows a lower entropy. With this, we can conclude that having a dynamic weight between re-observation and exploration based on the distance from the frontier point helps in keeping a better map but affects slightly the exploration speed. In general, while both approaches seem beneficial, using $u_2$ (A\_O) seems to give them more advantages. Note also that, in the current implementation, the visibility range is computed on the 2D domain to keep the computation feasible. Therefore, we are `discarding' a lot of information since many flat surfaces or cells behind low objects can potentially be seen from any given point of view.

The objective of A\_L is to show that our approach has no issues in completely exploring the environment while keeping the path shorter and the same final normalized entropy. Moreover, we also find that if run for a longer period of time, our approach further refines its results and reaches better metrics overall.

\begin{figure*}[!htbp]
    \centering
    \subfloat[Path length evolution]{
    \includegraphics[width=.8\columnwidth]{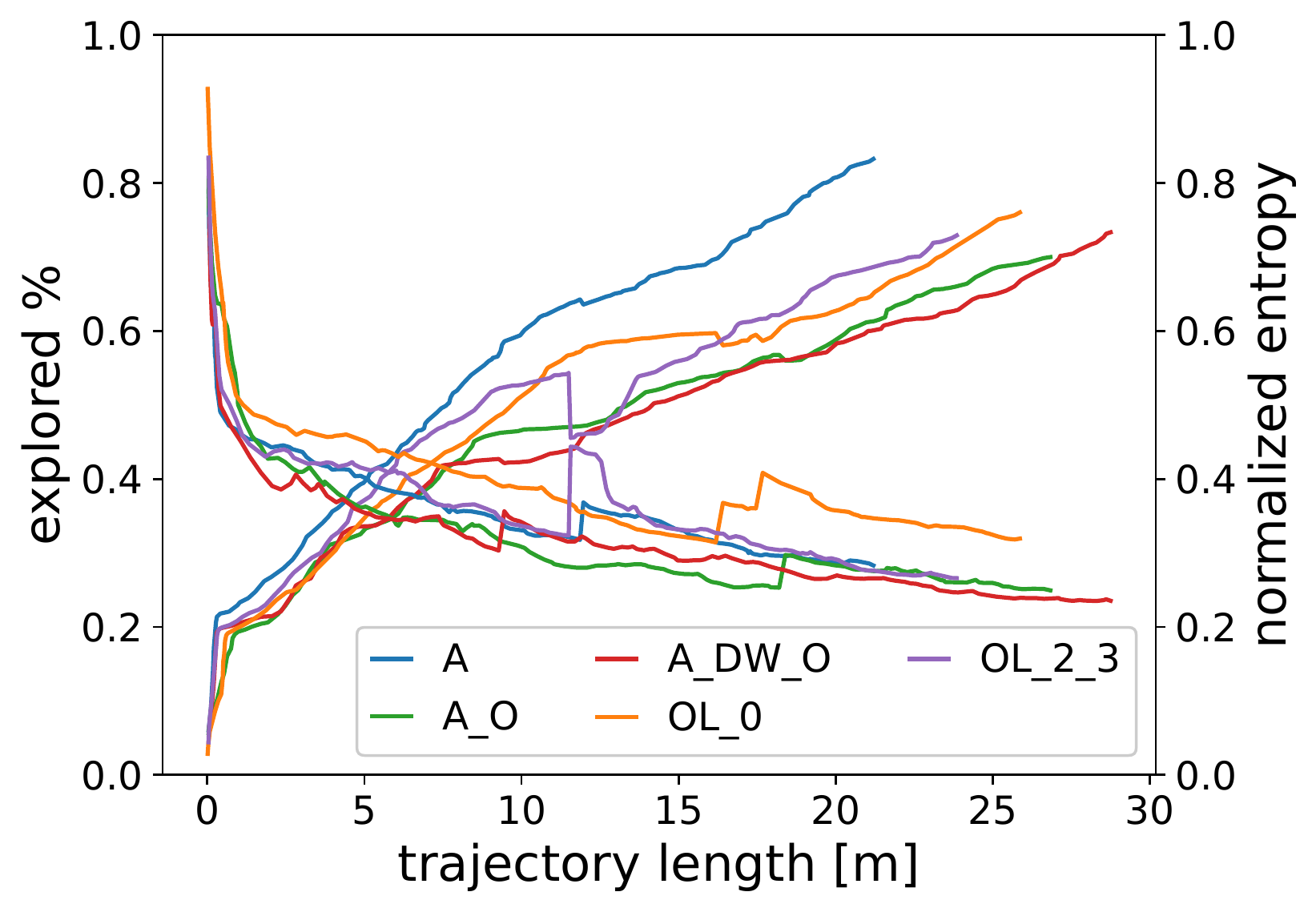}
    }
    \subfloat[Time evolution]{\includegraphics[width=.8\columnwidth]{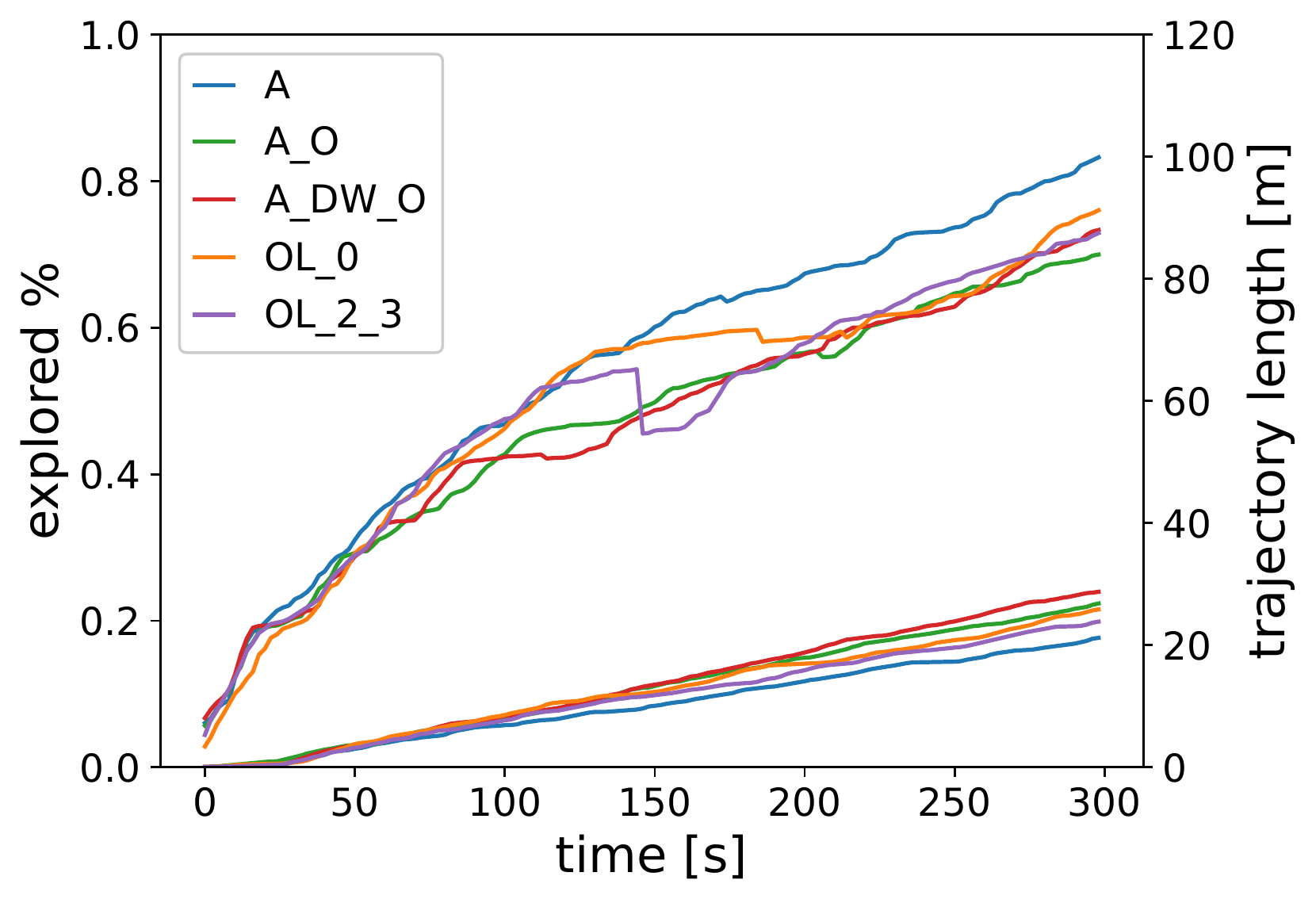}
    }\caption{Comparison of our complete approach with the state-of-the-art methods on our real robot in an office reception area.}
    \label{fig:realexpresults}
\end{figure*}

Finally, the slightly lower final explored area of our method w.r.t. OL\_2\_3 is explained by two main factors. Foremost, the computations of the first level of activeness cause a natural overhead that, despite being reduced by the frustum overlap, is still significant. An improvement in this would be considering an approximated raycasting technique rather than a complete one. Secondly, both the third level of activeness and the computation of the next waypoint optimal-heading cause a slight start and stop behaviour. This can be clearly seen in both~\fig{fig:gazeolact} and~\fig{fig:awsolact} when comparing OL\_* with OL\_*\_3. Anyway, a finer tuning of the parameters should solve this problem.

\begin{figure*}[!ht]
    \centering
    \subfloat[OL\_0 - Café]{\includegraphics[trim={2cm 2cm 2cm 2cm},clip,width=.42\columnwidth]{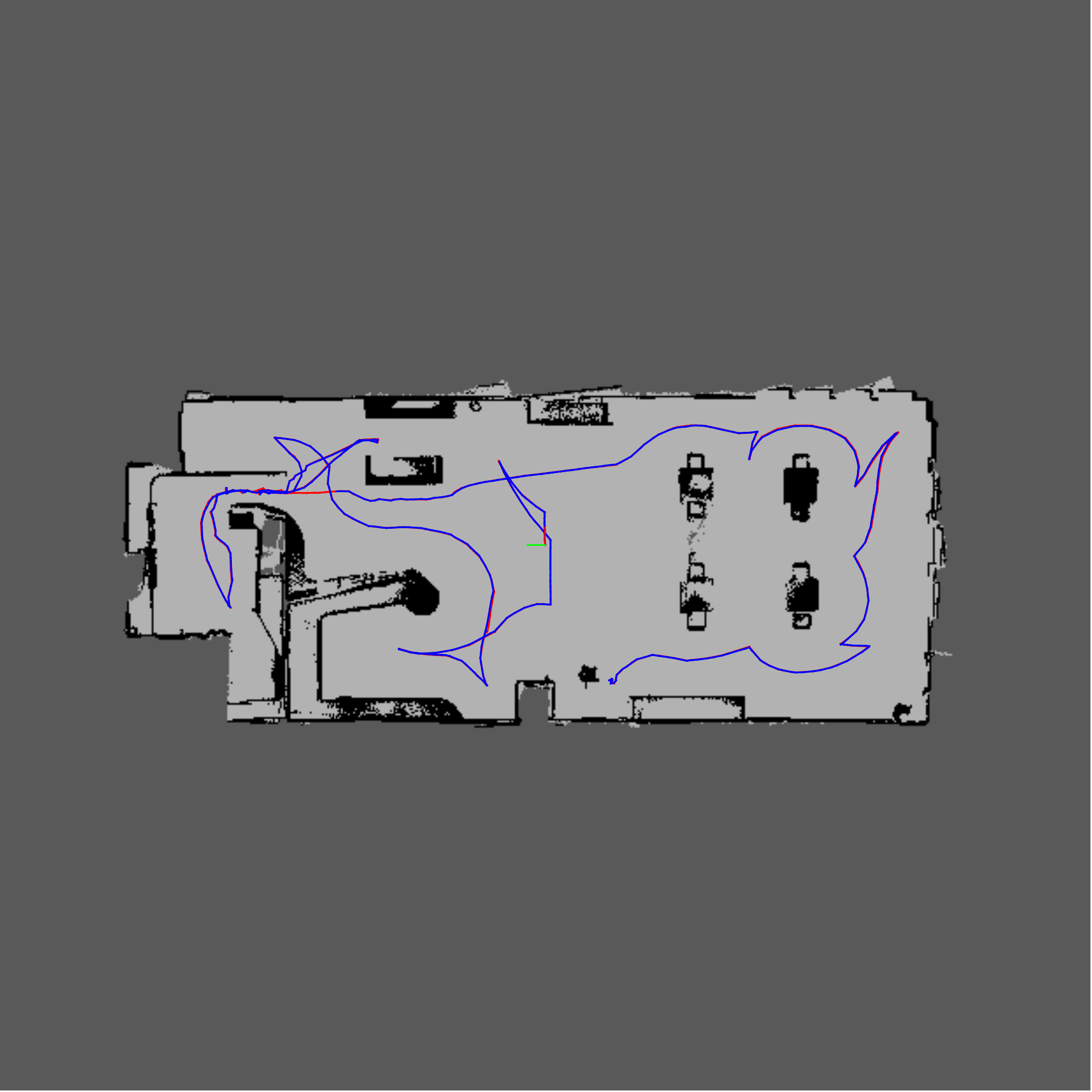}}  
    \subfloat[OL\_2\_3 - Café]{\includegraphics[trim={2cm 2cm 2cm 2cm},clip,width=.42\columnwidth]{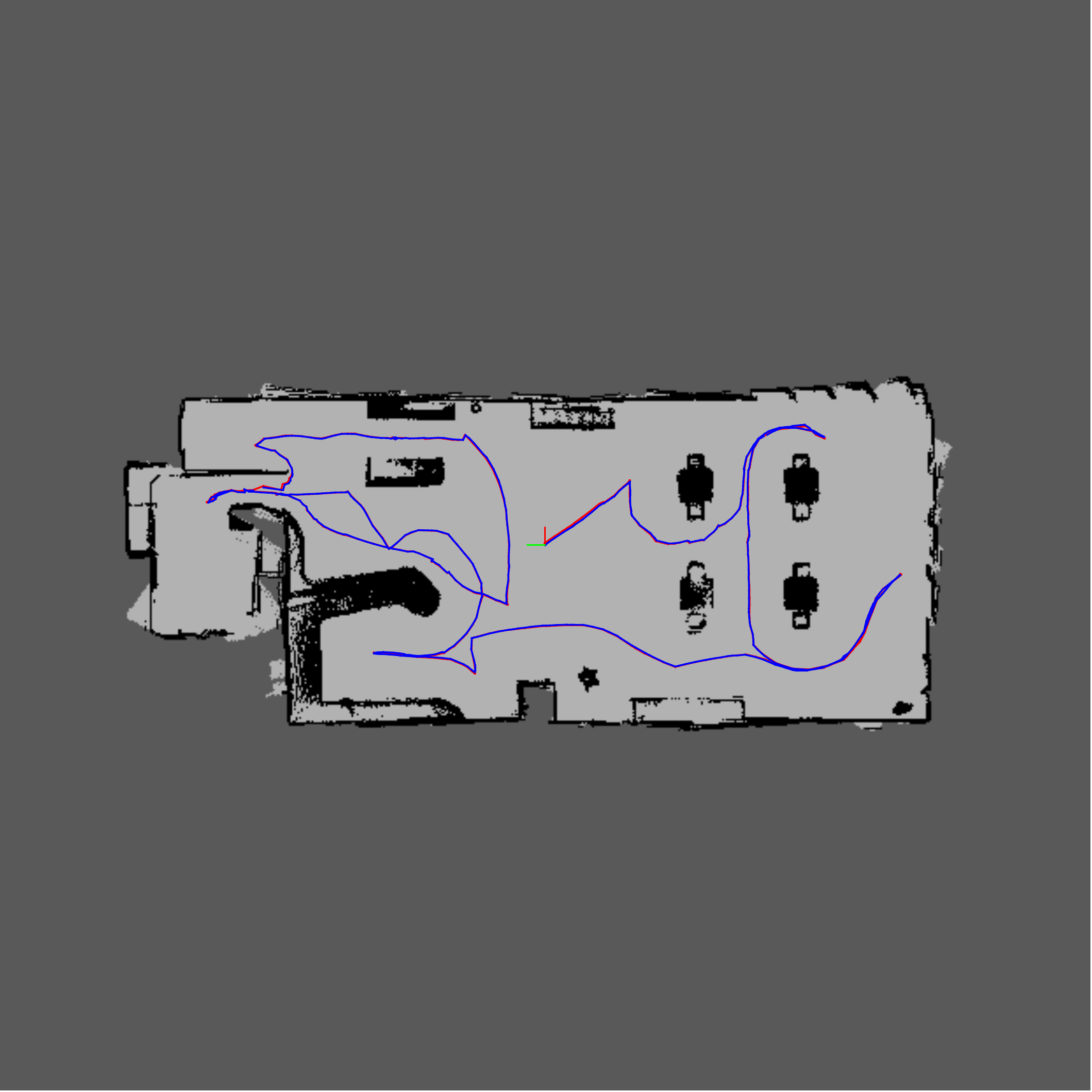}}
    \subfloat[A - Café]{\includegraphics[trim={2cm 2cm 2cm 2cm},clip,width=.42\columnwidth]{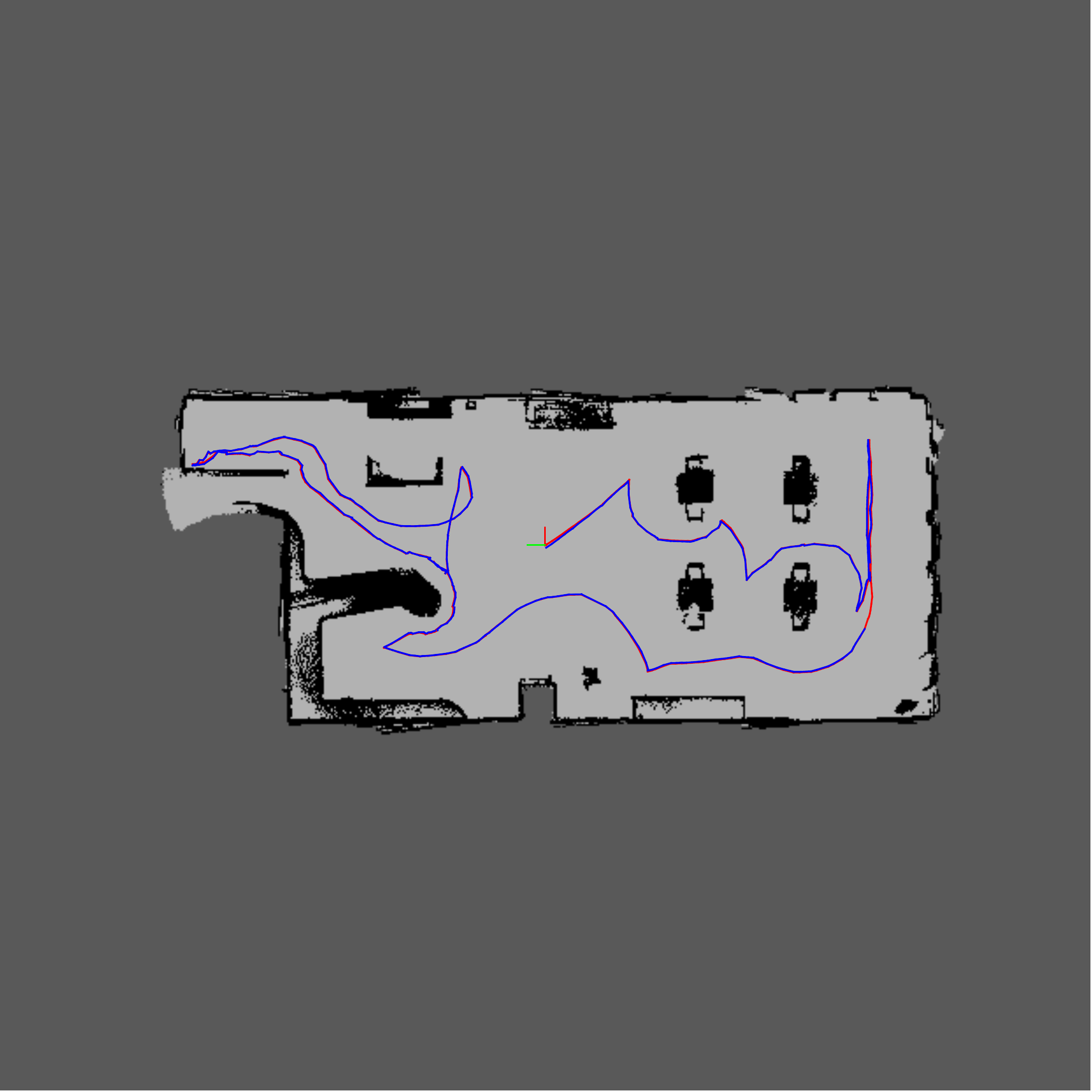}}
    \subfloat[A\_DW\_O - Café]{\includegraphics[trim={2cm 2cm 2cm 2cm},clip,width=.42\columnwidth]{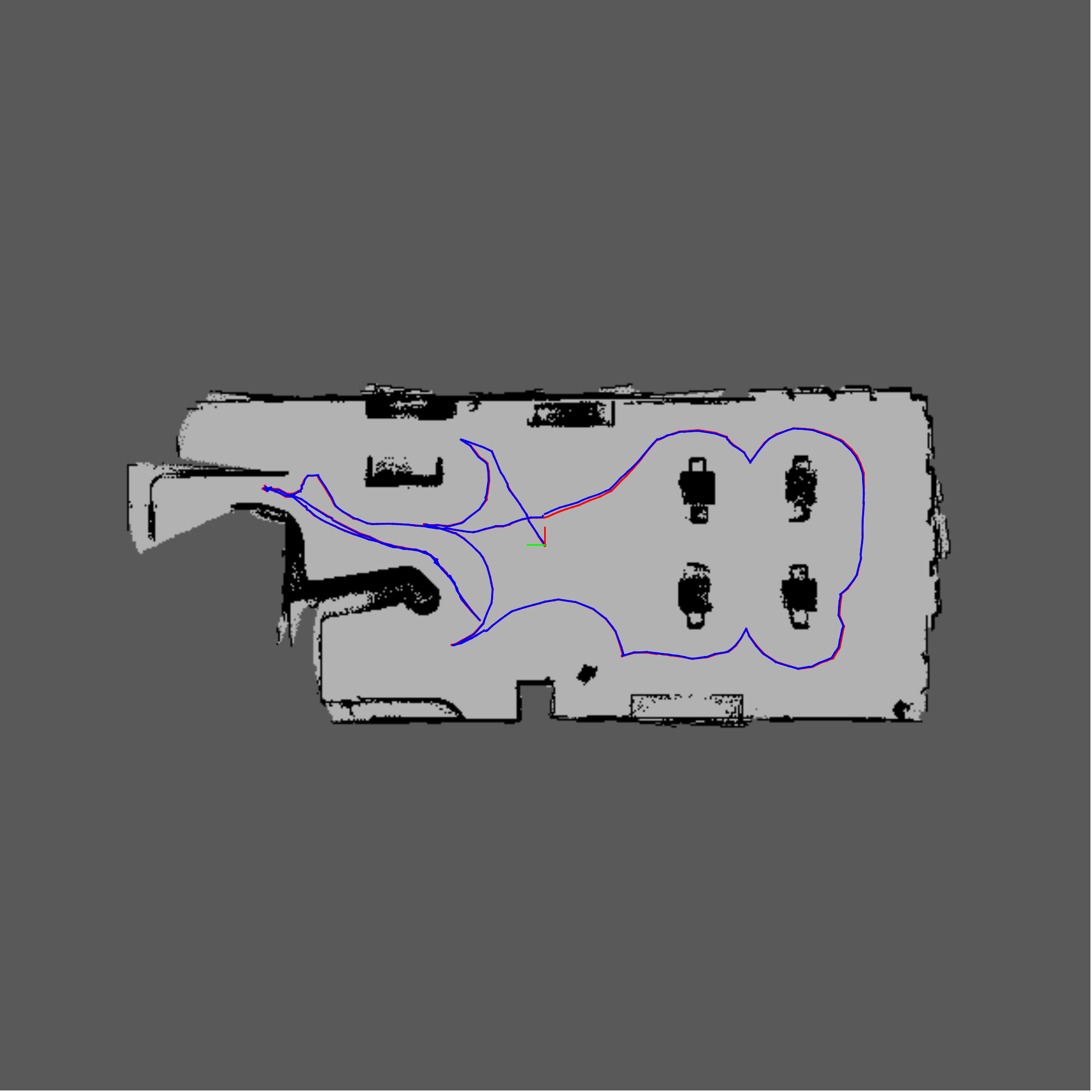}}
    \subfloat[Groundtruth - Café]{\includegraphics[trim={1.5cm 1.5cm 1.5cm 1.5cm},clip,width=.42\columnwidth]{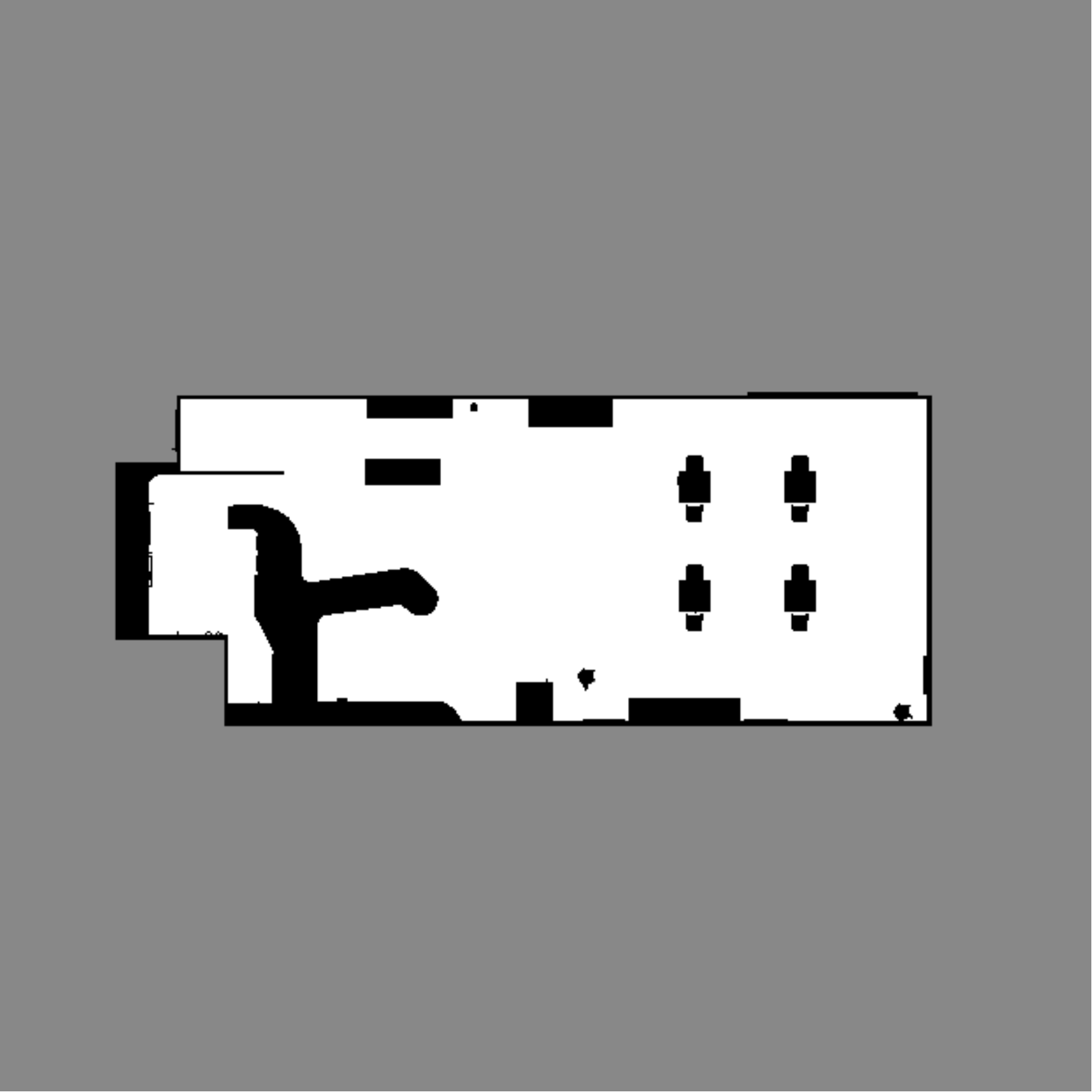}}
    
    \subfloat[OL\_0 - Small house]{\includegraphics[trim={2cm 2cm 2cm 2cm},clip,width=.42\columnwidth]{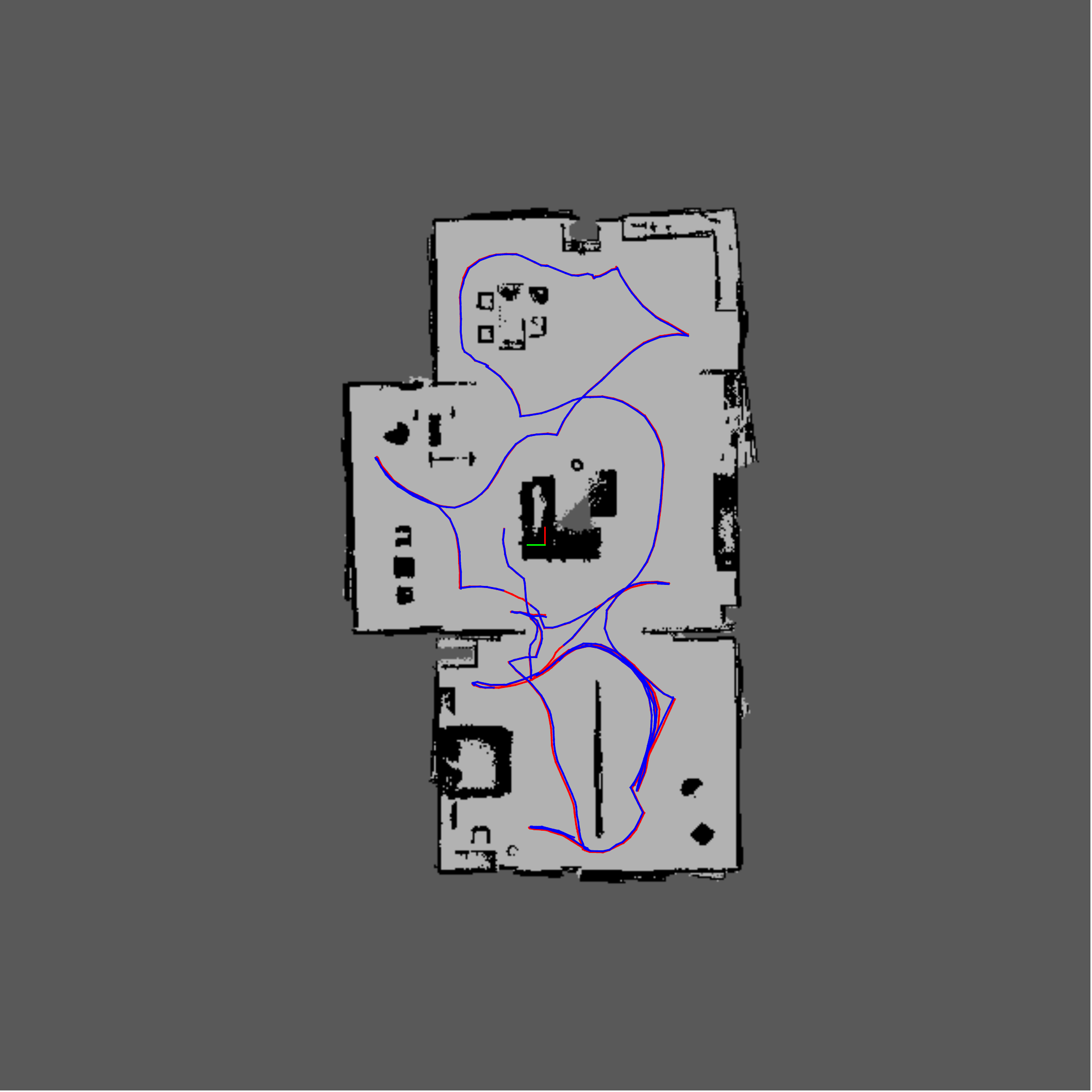}}
    \subfloat[OL\_2\_3 - Small house]{\includegraphics[trim={2cm 2cm 2cm 2cm},clip,width=.42\columnwidth]{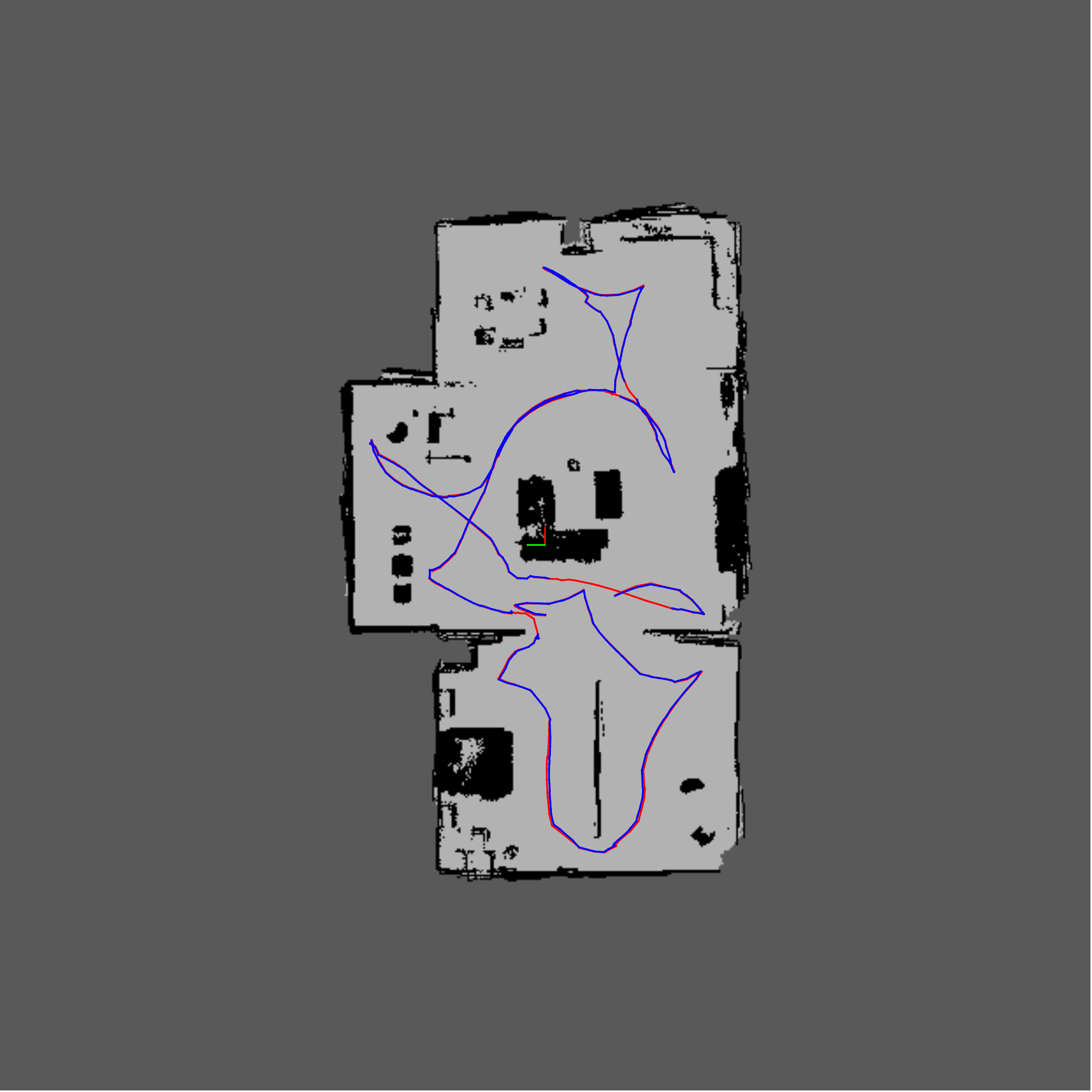}}
    \subfloat[A - Small house]{\includegraphics[trim={2cm 2cm 2cm 2cm},clip,width=.42\columnwidth]{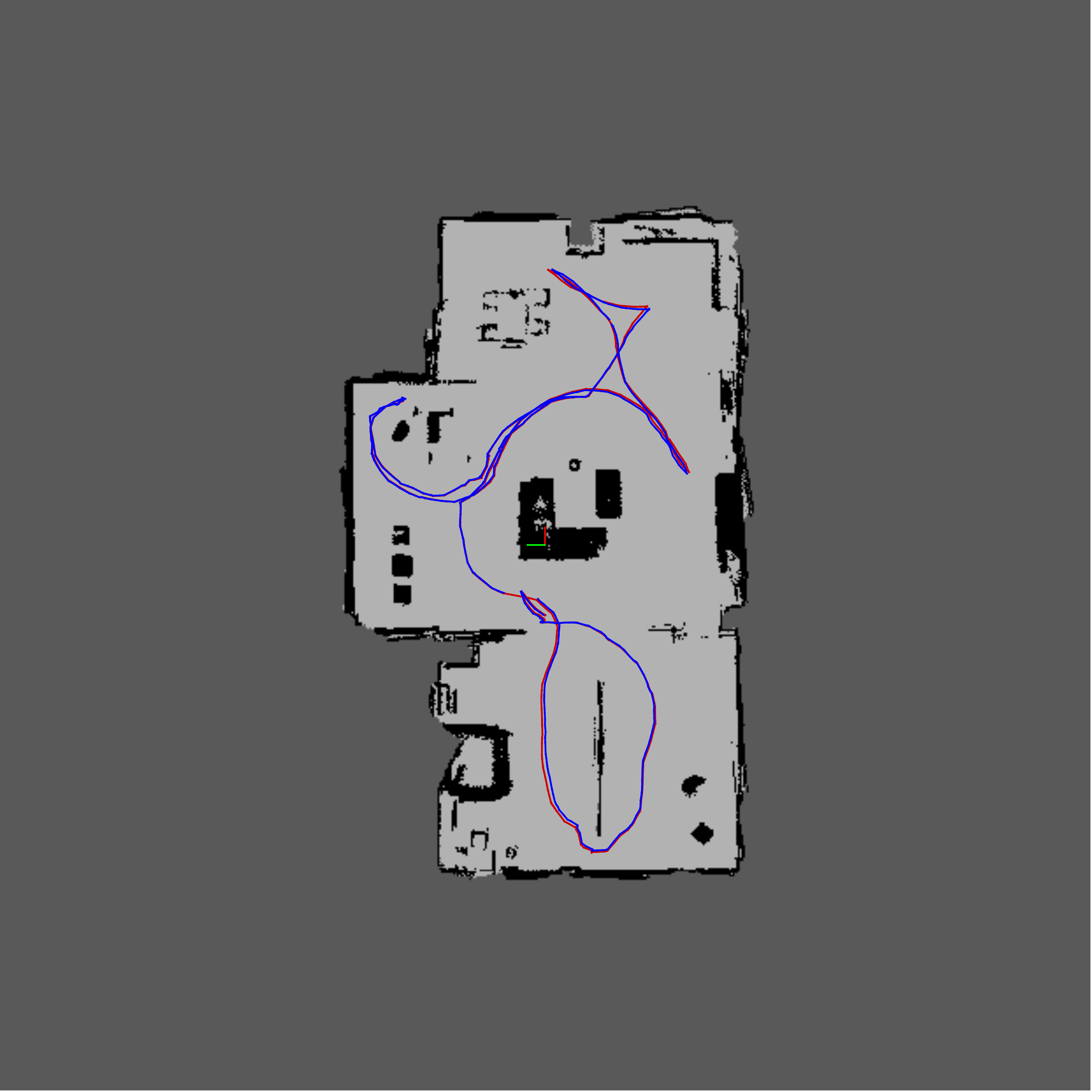}}
    \subfloat[A\_DW\_O - Small house]{\includegraphics[trim={2cm 2cm 2cm 2cm},clip,width=.42\columnwidth]{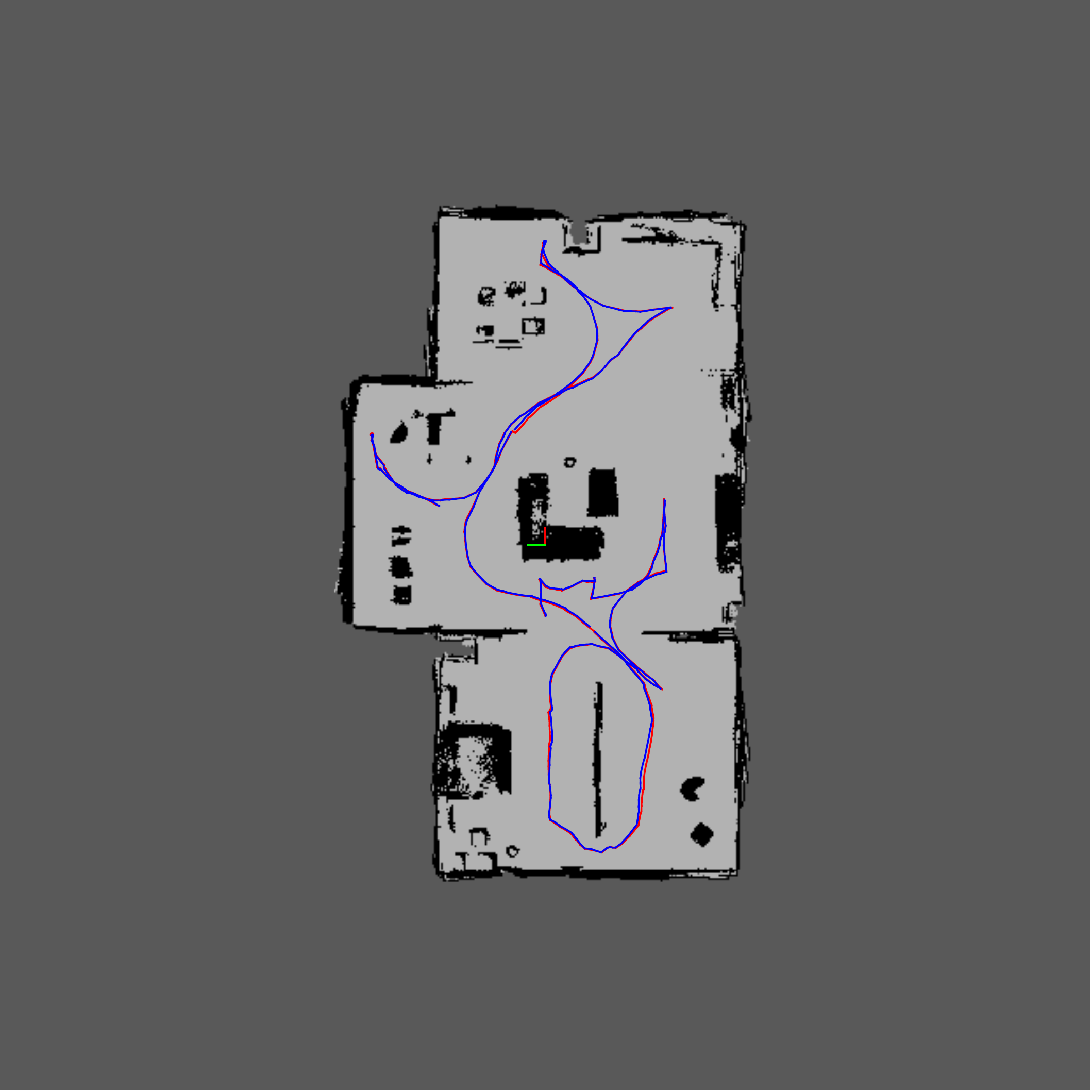}}
    \subfloat[Groundtruth - Small house]{\includegraphics[trim={2cm 2cm 2cm 2cm},clip,width=.42\columnwidth]{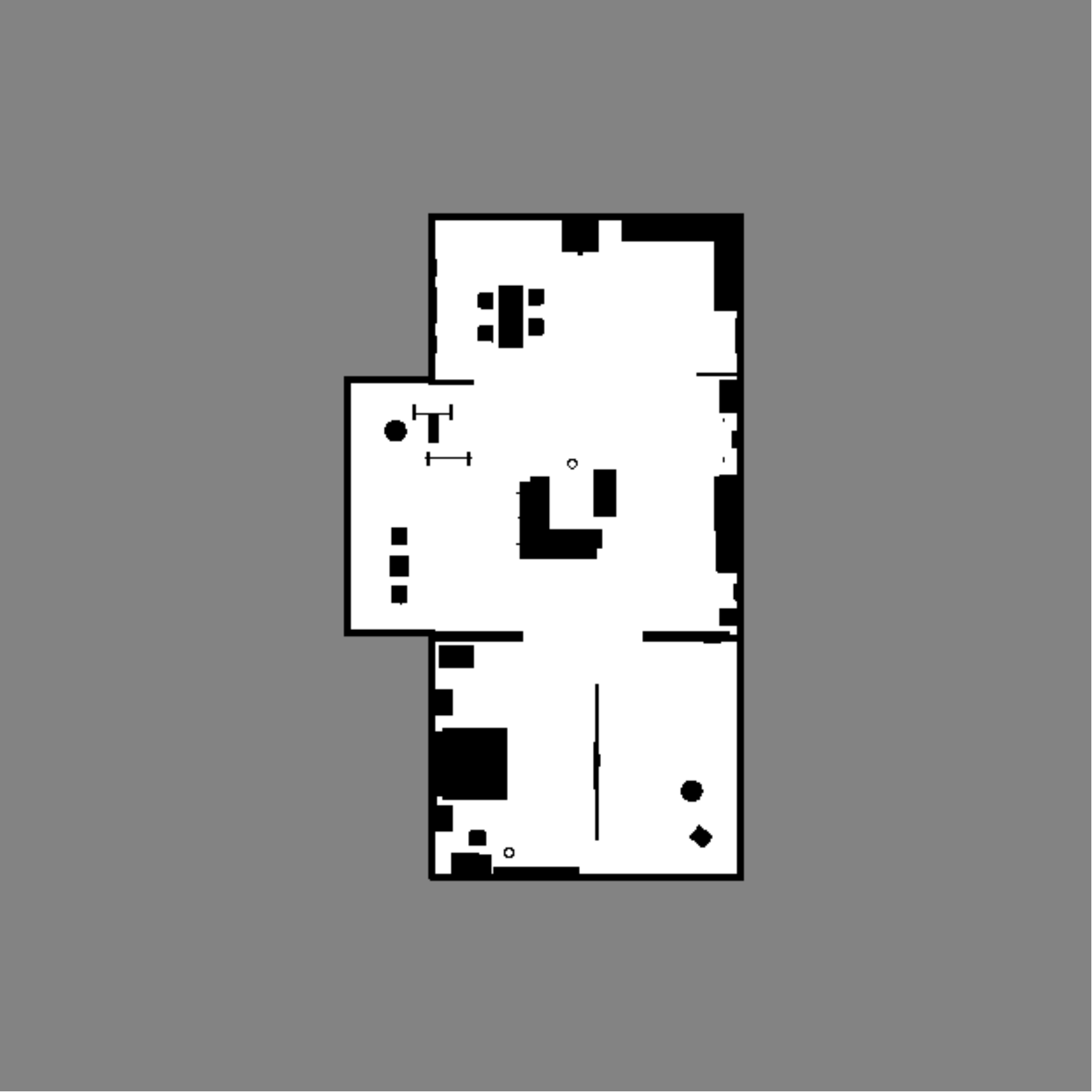}}
    
    \caption{Examples of grid maps with overlaid pose graphs for analyzed methods OL\_0, OL\_2\_3, A, A\_DW\_O and groundtruth. In blue the connections between nodes, in red the ground truth trajectory. The images were generated with the \textit{rtabmap-databaseViewer} tool and chosen to be representative. Loop closures are hidden for visibility sake.}
    \label{fig:examples}
\end{figure*}

\subsection{Real Robot Experiments}

The experiments with our real robot consist of five runs of five minutes each in a custom environment depicted in Fig.~\ref{fig:cover}. For safety reasons, the speed of the robot has been limited to $0.25$ m/s. For this set of experiments, the ground truth of both the robot position and the map was not available. Therefore, the analysis can only be done with respect to the area explored, the distance travelled, the final map entropy, and the number of loop closures as shown in Fig.~\ref{fig:realexpresults} and Tab.~\ref{tab:realexpresults}. 

The real robot results in general have the same trend as our simulation experiments, further validating our approach. 

Most importantly, as seen in Fig.~\ref{fig:realexpresults}, our complete approach A achieves the shortest path length and explores the maximum area while keeping a fairly low entropy. 

Both A\_O and A\_DW\_O, i.e. the methods using our new proposed utilities, are the best ones considering the final normalized entropy and the average loop closures. 

Furthermore, we can see in table Tab.~\ref{tab:realexpresults} that OL\_0 achieves more loop closures than OL\_2\_3, despite having a much higher normalized entropy. This is slightly in contrast to the simulation results. 
\begin{table}[!ht]
\centering
\resizebox{\columnwidth}{!}{
\begin{tabular}{lcccccc}
\hline
\multicolumn{1}{c}{} & \multicolumn{2}{c}{Area}                             & \multicolumn{2}{c}{Normalized Entropy} & \multicolumn{2}{c}{Loops per m} \\ \hline
                         & mean {[}$\mathrm{m}^2${]} & std {[}$\mathrm{m}^2${]} & mean               & std               & mean           & std            \\ \hline
A            & $83.257$                  & $13.239$                 & $0.283$            & $0.054$           & $2.228$        & $1.488$        \\
A\_O                  & $69.993$                  & $5.205$                  & $0.249$            & $0.042$           & $3.852$        & $1.434$        \\
A\_DW\_O              & $73.344$                  & $1.371$                  & $0.235$            & $0.015$           & $3.051$        & $1.611$        \\
OL\_0                 & $76.037$                  & $5.774$                  & $0.319$            & $0.101$           & $2.397$        & $1.057$        \\
OL\_2\_3              & $72.924$                  & $18.419$                 & $0.266$            & $0.041$           & $1.016$        & $0.384$
\end{tabular}}
\caption{Real robot results: comparison of proposed methods and utilities using various metrics.}
\label{tab:realexpresults}
\end{table}
Overall, even if we recognize that the real robot experiments are not as statistically significant as the simulation ones, they clearly indicate that our three levels of activeness improve the final results bringing benefits to all the comparison metrics. 

Lastly, it must be noted that we get longer paths with  A\_O and A\_DW\_O. This is most likely linked to the very low velocity of the robot. We noticed that while these two methods facilitate longer and more informative paths, the OL methods usually facilitate the nearest frontier. In A\_O and A\_DW\_O methods, the robot suffers from several start-and-stop instances and the continuous necessity to re-computed goals. 

\begin{figure*}[!ht]
    \centering 
    \subfloat[OL\_0 - Real world]{\includegraphics[trim={3cm 8cm 7cm 2cm},clip,width=.51\columnwidth]{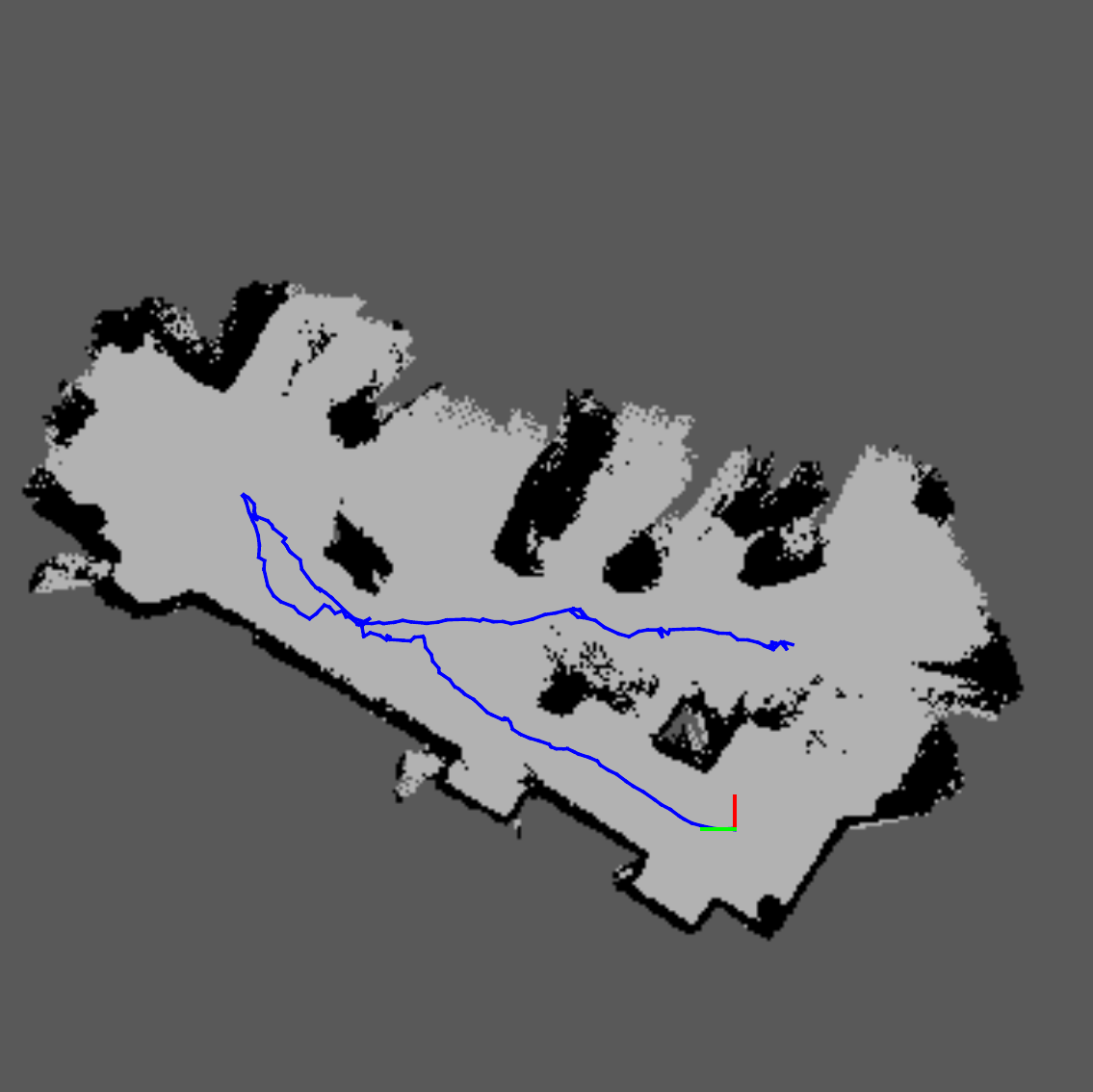}}
    \subfloat[OL\_2\_3 - Real world]{\includegraphics[trim={4.5cm 8cm 5.5cm 2cm},clip,width=.51\columnwidth]{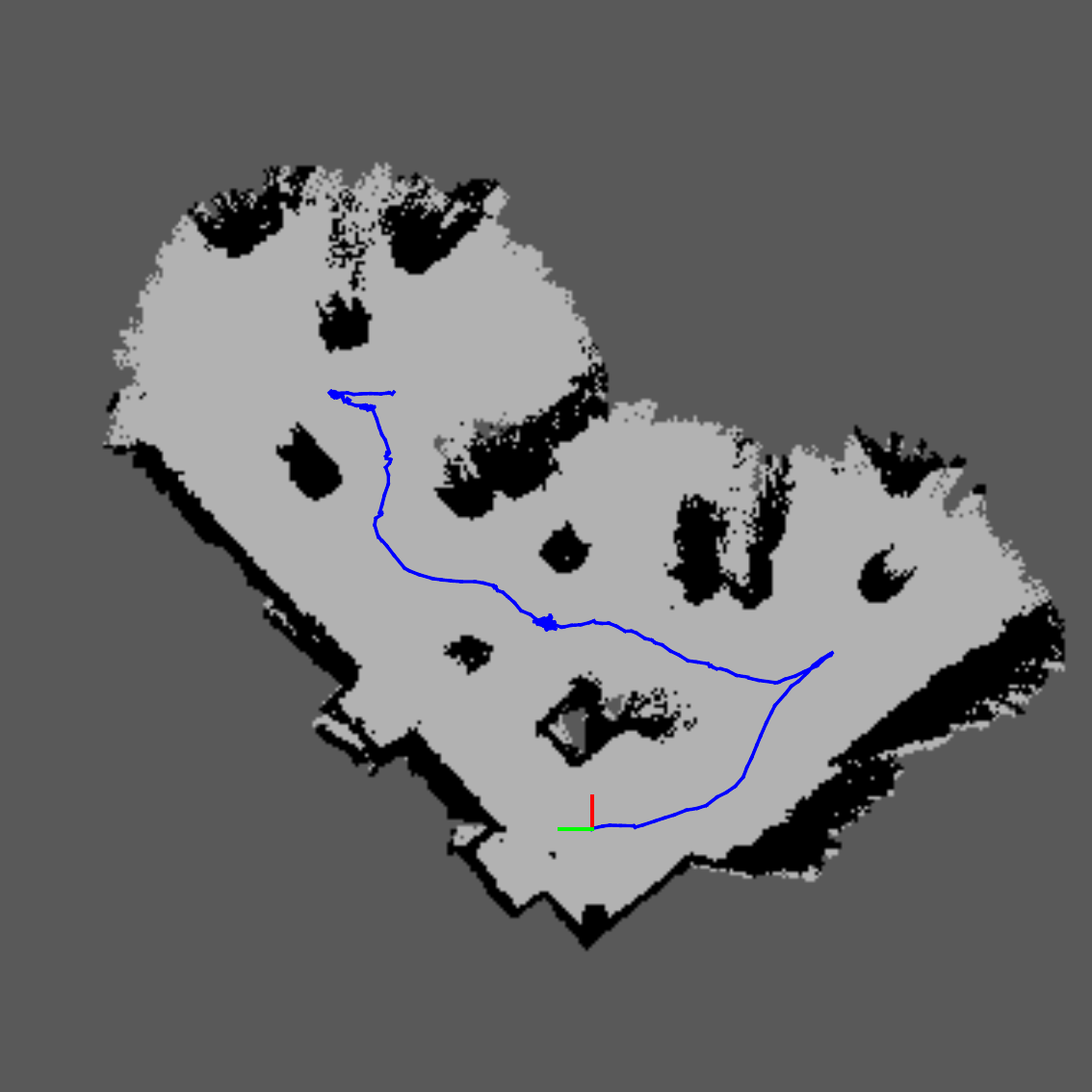}}
    \subfloat[A - Real world]{\includegraphics[trim={3cm 8cm 7cm 2cm},clip,width=.51\columnwidth]{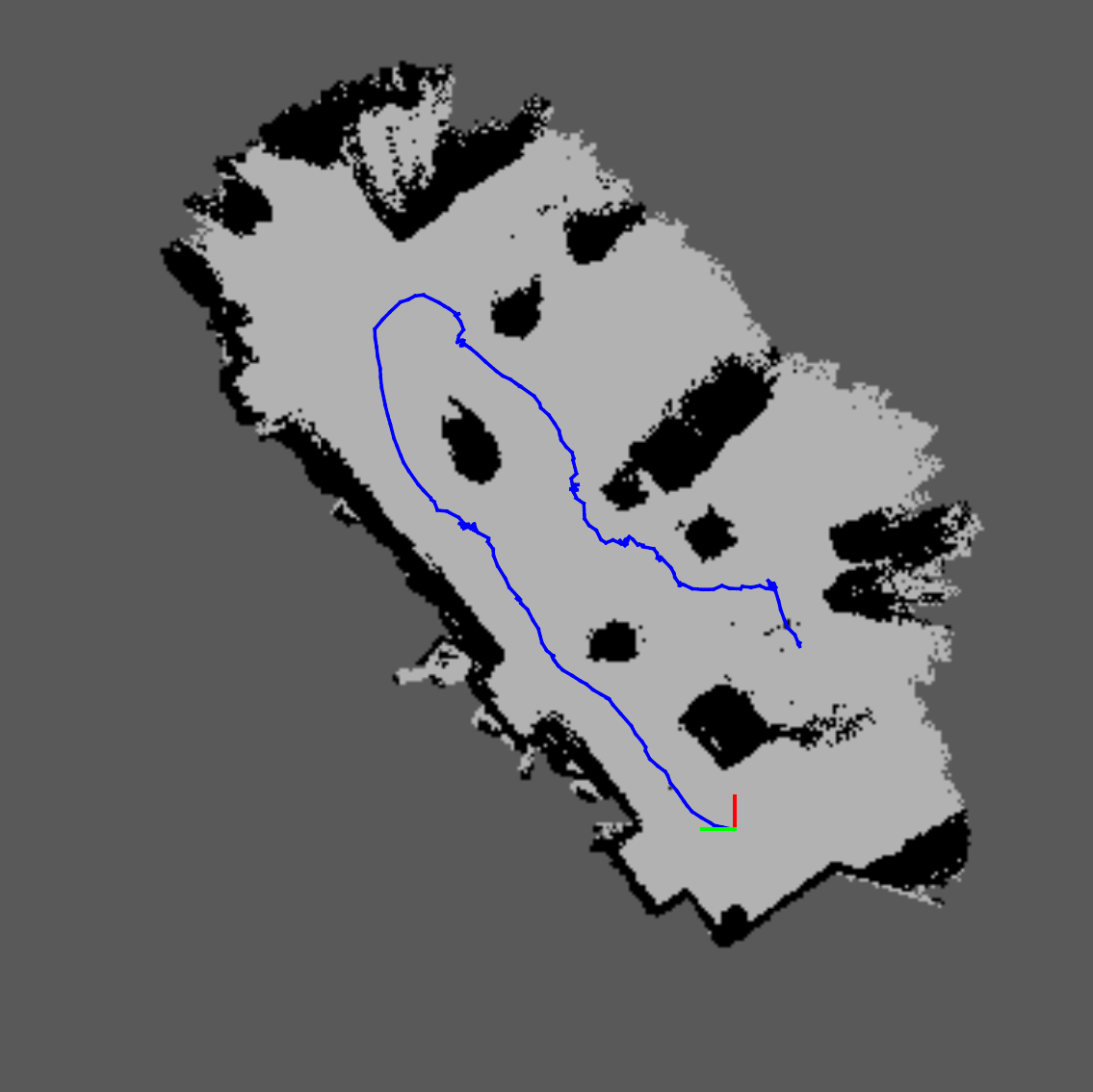}}
    \subfloat[A\_DW\_O - Real world]{\includegraphics[trim={5cm 8cm 5cm 2cm},clip,width=.51\columnwidth]{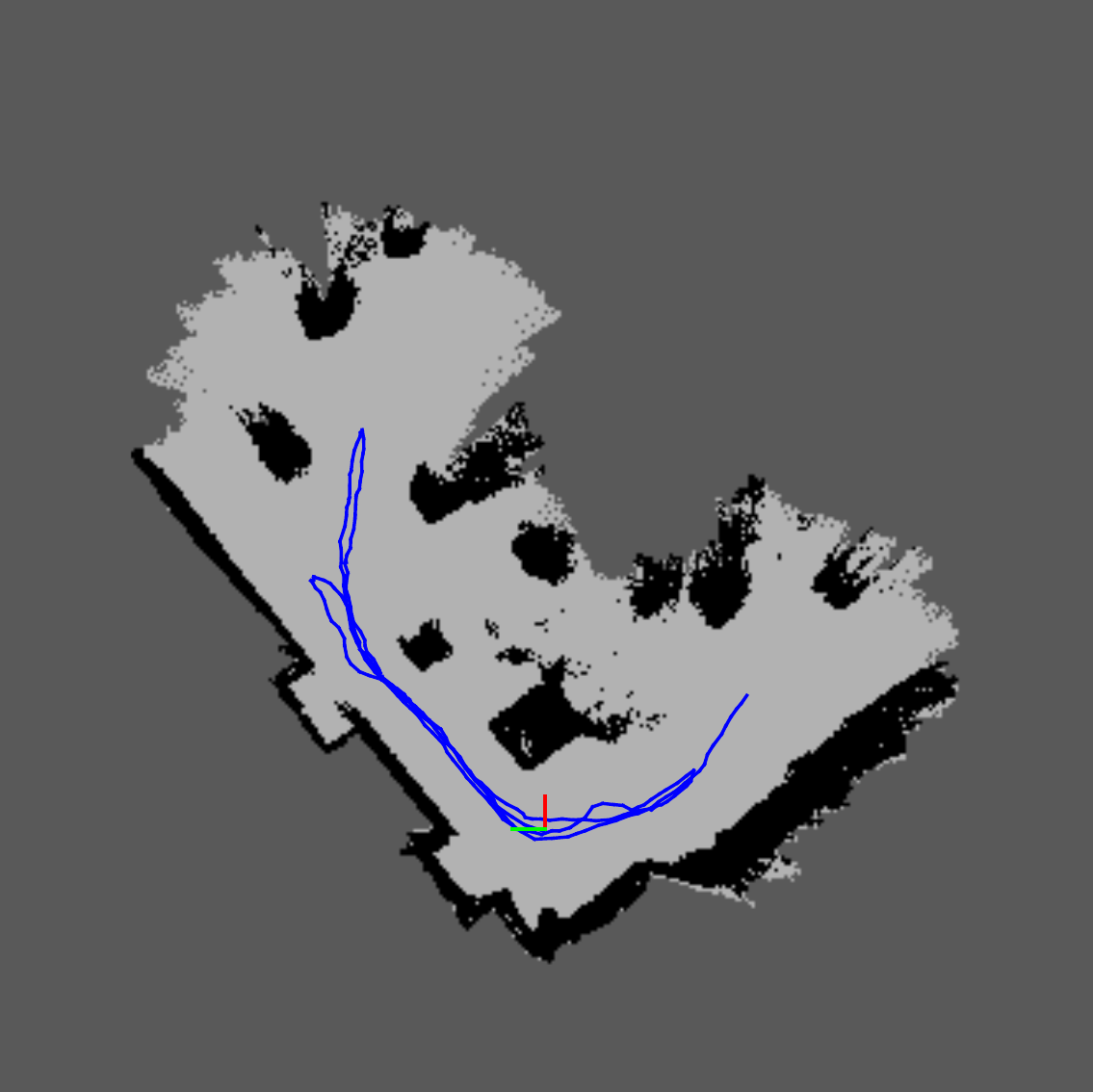}}

    \caption{Examples of grid maps with overlaid pose graphs for analyzed methods OL\_0, OL\_2\_3, A, A\_DW\_O in the real world settings. The images were generated with the \textit{rtabmap-databaseViewer} tool and chosen to be representative. In blue the connections between nodes. \textit{Recall:} groundtruth was not available. Loop closures hidden for visibility sake.}
    \label{fig:examples_real}
\end{figure*}

\subsection{Further remarks}
For both simulation and real robot experiments, one might notice drops in the exploration amount and spikes in the normalized entropy. Those are related to two main factors. The first one is related to the RTABMap system itself since, once a loop closure is triggered, the whole map gets rectified and updated with subsequent possible degradation and corruption. The second one is related to the recovery procedure. In this case, the current explored area and its normalized entropy are like the ones at the beginning of the experiment, until a loop closure with the previous session is found. We do not use a cumulative sum of the explored area among the two sessions since we have no way to ensure that the `newly' explored space is different w.r.t. the previous one. Once a loop closure is found with the previous map area the two get automatically merged, `restoring' these two values.
	
\section{CONCLUSIONS}
\label{sec:conc}

In this work, we presented a novel active V-SLAM method for omnidirectional robots. Our novel approach consists of the combination of three levels of activeness. In the first level, the robot selects goals and decides on the most informative paths to them. The second level re-optimizes waypoints along the path in a way such that the real-time updated map information is continuously exploited. The third level of activeness ensures that the robot orientations have maximum visual feature visibility. This eventually ensures better localization accuracy, in addition to lower map entropy. The most significant result of our approach is that it requires up to 39\% less path traversal to obtain similarly good coverage and lower map entropy compared to methods that are tuned versions of the state-of-the-art approaches~\cite{CarrilloDamesKumarCastellanos2017}. We demonstrated the efficacy of our method both in simulation and in real robot scenarios. A drawback of our method is that the robot encounters some start-stop situations in which the local paths are replanned. While this may not be desirable in some applications, one way to avoid such a behaviour is to employ a better computational unit.
\
Future direction of our work includes active V-SLAM with moving objects and persons in the environment, and extension of our approach to aerial robots.

    \section*{ACKNOWLEDGMENT}{The authors would like to thank Heiko Ott and Mason Landry for their help with the design of the hardware. \\
    
        The authors thank the International Max Planck Research School for Intelligent Systems (IMPRS-IS) for supporting Elia Bonetto and Pascal Goldschmid.}
%
	
	\bibliographystyle{elsarticle-num}

\end{document}